\documentclass[10pt,twocolumn,letterpaper]{article}

\usepackage{cvpr}
\usepackage{times}
\usepackage{epsfig}
\usepackage{graphicx}
\usepackage{amsmath}
\usepackage{amssymb}

\usepackage{eso-pic}
\usepackage{subfig} % added
\usepackage{booktabs}   % added
\usepackage{multirow}   % added
\usepackage[bottom]{footmisc}
\usepackage{nopageno}
\usepackage{lipsum}

\usepackage{kotex}

\newcommand*{\affaddr}[1]{#1} % No op here. Customize it for different styles.
\newcommand*{\affmark}[1][*]{\textsuperscript{#1}}
\newcommand*{\email}[1]{\texttt{#1}}

\newcommand\blfootnote[1]{%
  \begingroup
  \renewcommand\thefootnote{}\footnote{#1}%
  \addtocounter{footnote}{-1}%
  \endgroup
}

\newenvironment{myindentpar}[1]%
  {\begin{list}{}%
          {\setlength{\leftmargin}{#1}}%
          \item[]%
  }
  {\end{list}}

\usepackage[pagebackref=true,breaklinks=true,letterpaper=true,colorlinks,bookmarks=false]{hyperref}
\cvprfinalcopy % *** Uncomment this line for the final submission

\def\cvprPaperID{4803} % *** Enter the CVPR Paper ID here

\ifcvprfinal\fi
\begin{document}

%%%%%%%%% TITLE
\title{Reference-Based Sketch Image Colorization using\\ Augmented-Self Reference and Dense Semantic Correspondence}

\author{Junsoo Lee\affmark[*,1],\quad Eungyeup Kim\affmark[*,1],\quad Yunsung Lee\affmark[2],\quad Dongjun Kim\affmark[1],\quad Jaehyuk Chang\affmark[3],\quad Jaegul Choo\affmark[1] \\
\affaddr{\affmark[1]KAIST,\quad \affmark[2]Korea University,\quad \affmark[3]NAVER WEBTOON Corp.} \\
\small{\email{\{junsoolee93,eykim94,rassilon,jchoo\}@kaist.ac.kr}}, \\ \small{\email{swack9751@korea.ac.kr},\quad \email{jaehyuk.chang@webtoonscorp.com}}
% For a paper whose authors are all at the same institution,
% omit the following lines up until the closing ``}''.
% Additional authors and addresses can be added with ``\and'',
% just like the second author.
% To save space, use either the email address or home page, not both
% \and
% Second Author\\
% Institution2\\
% First line of institution2 address\\
% {\tt\small secondauthor@i2.org}
}

\twocolumn[{%
\renewcommand\twocolumn[1][]{#1}%
\maketitle
\begin{center}
    \centering
    \includegraphics[width=\linewidth]{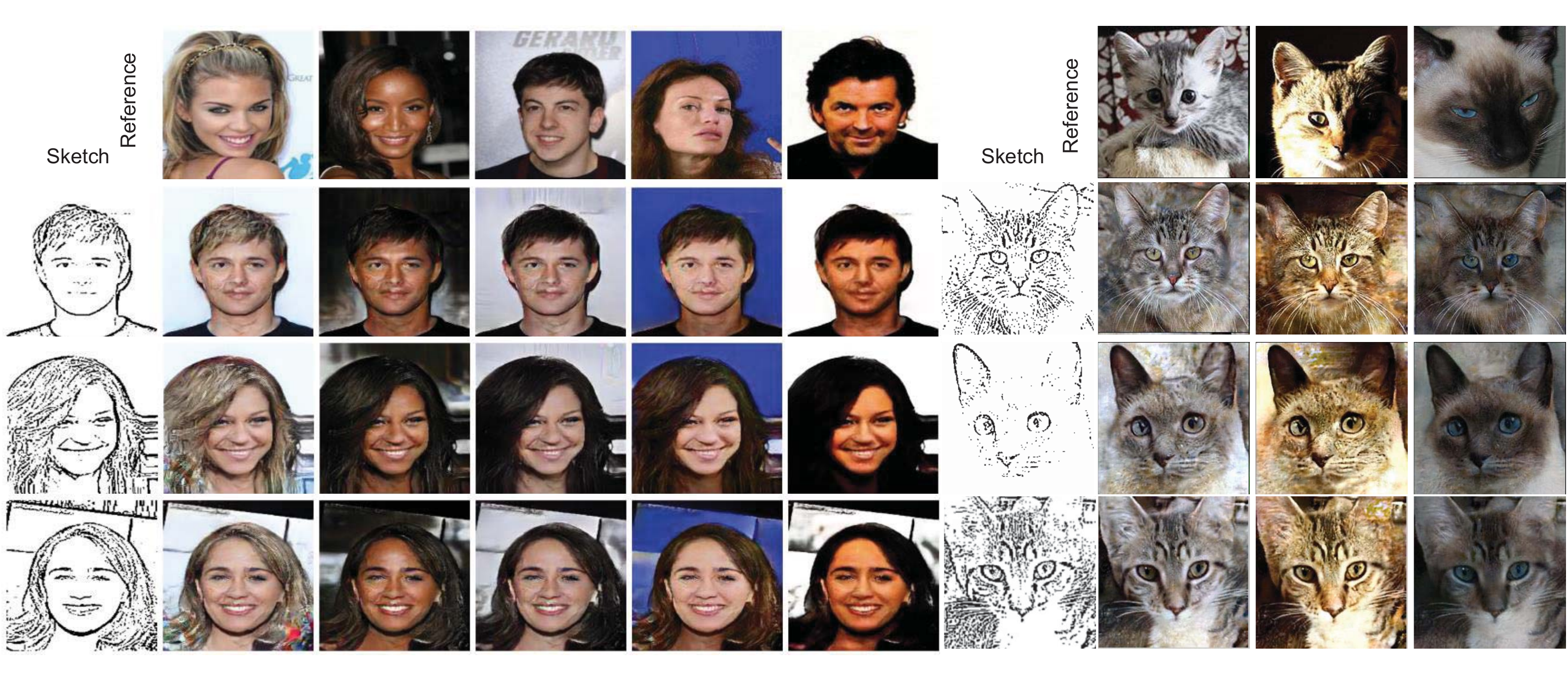}
    \captionof{figure}{Qualitative results of our method on the CelebA~\cite{liu2015celeba} and ImageNet~\cite{russakovsky2015imagenet} dataset respectively. Each row has the same content while each column has the same reference.}
    % \caption{Qualitative results of our method on the CelebA dataset. }
    \label{fig:our_celeba2}
\end{center}
}]
\setlength{\skip\footins}{2mm}
\blfootnote{* indicates equal contribution}
\maketitle
%\thispagestyle{empty}
% \footnote{* example}

%%%%%%%%% ABSTRACT
\begin{abstract}
This paper tackles the automatic colorization task of a sketch image given an already-colored reference image. Colorizing a sketch image is in high demand in comics, animation, and other content creation applications, but it suffers from information scarcity of a sketch image. To address this, a reference image can render the colorization process in a reliable and user-driven manner. However, it is difficult to prepare for a training data set that has a sufficient amount of semantically meaningful pairs of images as well as the ground truth for a colored image reflecting a given reference (e.g., coloring a sketch of an originally blue car given a reference green car). To tackle this challenge, we propose to utilize the identical image with geometric distortion as a virtual reference, which makes it possible to secure the ground truth for a colored output image. Furthermore, it naturally provides the ground truth for dense semantic correspondence, which we utilize in our internal attention mechanism for color transfer from reference to sketch input. We demonstrate the effectiveness of our approach in various types of sketch image colorization via quantitative as well as qualitative evaluation against existing methods. 
\end{abstract}

\vspace{-4.0mm}

%%%%%%%%% BODY TEXT
\section{Introduction}
 Early colorization tasks~\cite{zhang2016colorful, larsson2016learning, larsson2017colorization} have focused on colorizing a grayscale image, which have shown great progress so far. 
 More recently, the task of colorizing a given sketch or outline image has attracted a great deal of attention in both computer vision and graphics communities, due to its significant needs in practice. 
 Compared to a grayscale image, which still contains the pixel intensity, a sketch image is information-scarce, making its colorization challenging in nature. 
%  To remedy this issue, an already colored image, which shares the same semantic objects and involves various user intent, can be provided as a reference to make the colorization of a sketch image practically useful and meaningful. 
%  This task can be viewed as a type of conditional image colorization where a reference image works as a condition or hint. 
%  To remedy this issue, an approach of imposing additional conditions for the colorization task has been widely explored. 
%  This setting is practically useful because it helps not only improve the output quality but also produce different results according to the conditions provided by users. (->related work 2.2)
%  There are generally two types of methods for providing additional information: user hints and reference image. Details are described in Section~\ref{sec:conditional_image_colorization}.
 To remedy this issue, generally two types of approach of imposing additional conditions to the sketch image have been explored: user hints and reference image. % (Section~\ref{sec:conditional_image_colorization}).
 % These settings are practically useful because they help not only improve the output quality but also produce different results according to the conditions provided by users.
 % Details are described in Section~\ref{sec:conditional_image_colorization}.

 % In the first form, manual interactions, e.g., scribbles or points, are directly provided as hints for synthesizing the output. 
 % Since this setting requires human interactions, which contain positions as well as the palette of color, it has some limitation of expanding automatic colorization on the large scales at test time.
%  As explained in Section~\ref{sec:conditional_image_colorization}, there are previous works utilizing a reference or already-colored image, which shares the same semantic object of the target image. 
%  It requires an ability for the model to establish visual correspondences and inject colors through the mappings from the reference to the target.  
%  Since there are few datasets containing the label of the correspondence in the real world, however, the cost of pairing reliable source-reference images can be a bottleneck to expand the task on wide range of domains. 
%  Moreover, although existing studies described in Section~\ref{sec:sketch_based_task} have dealt with sketch image in various tasks, the sketch colorization guided by the reference is still under-explored mainly due to the huge information discrepancy between the sketch and reference.
 
 As explained in Section~\ref{sec:conditional_image_colorization}, there are previous works utilizing a reference or already-colored image, which shares the same semantic object of the target image. 
 It requires an ability for the model to establish visual correspondences and inject colors through the mappings from the reference to the target.
 However, due to the huge information discrepancy between the sketch and reference, the sketch colorization guided by the reference is still under-explored compared to other sketch-based tasks (Section~\ref{sec:sketch_based_task}).
 Moreover, there are few datasets containing the labels of the correspondence between the two images, and the cost of generating a reliable matching of source and reference becomes a critical bottleneck for this task over a wide range of domains. 
 
 %이 문단은 2.2 내용과 거의 일치 혹은 포함되는 내용인것 같아서, 많이 줄여도 좋을거같아요.
 
 In this work, we utilize an augmented-self reference which is generated from the original image by both color perturbation and geometric distortion.
 This reference contains the most of the contents from original image itself, thereby providing a full information of correspondence for the sketch, which is also from the same original image.
 Afterward, our model explicitly transfers the contextual representations obtained from the reference into the spatially corresponding positions of the sketch by the attention-based pixel-wise feature transfer module, which we term the spatially corresponding feature transfer (SCFT) module.
 Integration of these two methods naturally reveals groundtruth spatial correspondence for directly supervising such an attention module via our similarity-based triplet loss.
 This direct supervision encourages the network to be fully optimized in an end-to-end manner from the scratch and does not require any manually-annotated labels of visual correspondence between source-reference pairs.
 Furthermore, we introduce an evaluation metric which measures how faithfully the model transfers the colors of the reference in the corresponding regions of sketch.
 
 Both qualitative and quantitative experiments indicate that our approach exhibits the state-of-the-art performance to date in the task of information-scarce, sketch colorization based on a reference image.
 These promising results strongly demonstrate its significant potentials in practical applications in a wide range of domains.

 \section{Related work}
 \subsection{Sketch-based Tasks}
 \label{sec:sketch_based_task}
 %맨 앞줄로 이동시키는게 어떨까요? 그래서 전체적인 흐름이: sketch는 practical하고 좋은 점이 이러저러 있어. 그래서 관련 연구들도 좀 있어. 그런데 이런점 때문에 challenging하고 under-explored 되어있어. + 그래서 우리 work은 딱 이러한 점을 이렇게 극복해서 연구했던 거야. 까지..
 Sketch roughly visualizes the appearances of a scene or object by a series of lines.
 % By virtue of its simplicity to draw and potentiality in practical applications, sketch has been utilized in several tasks including image retrieval~\cite{DBLP:journals/corr/abs-1807-11724}, sketch recognition~\cite{liu2019sketchgan}, sketch generation~\cite{chen2018sketchygan, lu2018image}, and image inpainting~\cite{nazeri2019edgeconnect}.
 Thanks to its simple, easy-to-draw, and easy-to-edit advantages, sketch has been utilized in several tasks including image retrieval~\cite{yelamarthi2018zeroshot}, sketch recognition~\cite{liu2019sketchgan}, sketch generation~\cite{chen2018sketchygan, lu2018image}, and image inpainting~\cite{nazeri2019edgeconnect}.
 However, due to the lack of texture and color information in sketch image, the research on sketch-based image colorization, especially reference-based colorization, is quite challenging and still under-explored. 

 %스케치는 몇개의 선들을 통해 물체나 장면의 모습에 대해 대략적으로 시각화한다. practically 그려내기 쉽고 ~한 장점 덕분에 이전의 많은 연구들에서 스케치를 사용했었어 such as ~~~. 그러나 스케치 베이스의 컬러라이제이션 줰에서는 많이 under-explored되었는데, 이는 ~한 단점 때문에 꽤 challenging하기 때문. 우리 웤은 스케치를 컬러라이즈 함으로써 기존 그레이스케일보다 

 \subsection{Conditional Image Colorization}
 \label{sec:conditional_image_colorization}
 The automatic colorization has a limitation that users cannot manipulate the output with their desired color.
 To tackle this, recent methods come up with the idea of colorizing images with condition of the color given by users, such as scribbles~\cite{sangkloy2017scribbler}, color palette~\cite{zhang2017real, style2paints, paintschainer}, or text tags~\cite{kim2019tag2pix}.
 Even though these approaches have shown the impressive results in terms of the multi-modal colorization, they unavoidably require both precise color information and the geometric hints provided by users for every step.
 
 To overcome the inconvenience, an alternative approach, which utilizes an already colored image as a reference, has been introduced. 
 Due to the absence of geometric correspondence at the input level, early studies~\cite{kekre2008color, bugeau2013variational, liu-2008-intrinsic, Chia:2011:SCI:2070781.2024190, gupta2012image, charpiat2008automatic} utilized low-level hand-crafted features to establish visual correspondence. 
 % Recent studies~\cite{he2018deep, zhang2019deep, sun2019adversarial} focus on building the train dataset, which should be composed of reliable source-reference pairs because it is difficult for model to be trained by using a ground truth image. (not understandable)
 % Although these works propose a method to automatically compose the trainset, i.e., using features extracted from the pre-trained networks~\cite{he2018deep, zhang2019deep} or color histogram~\cite{sun2019adversarial}, it can be a bottleneck in practice because these method tends to be sensitive to domains. 
 Recent studies~\cite{he2018deep, zhang2019deep, sun2019adversarial} compose the semantically close source-reference pairs by using features extracted from the pre-trained networks ~\cite{he2018deep, zhang2019deep} or color histogram~\cite{sun2019adversarial} and exploit them in their training.
 These pair composition techniques however tend to be sensitive to domains, thereby limit their capability in % limited range of dataset domains. 
 a specific dataset.
 
 Our work presents a novel training scheme to learn visual correspondence by generating augmented-self reference in the self-supervised manner at the training time, and then demonstrates it's scalability on various type of datasets.
 % scalability -> validatity

\begin{figure*}
\begin{center}
\includegraphics[width=\linewidth]{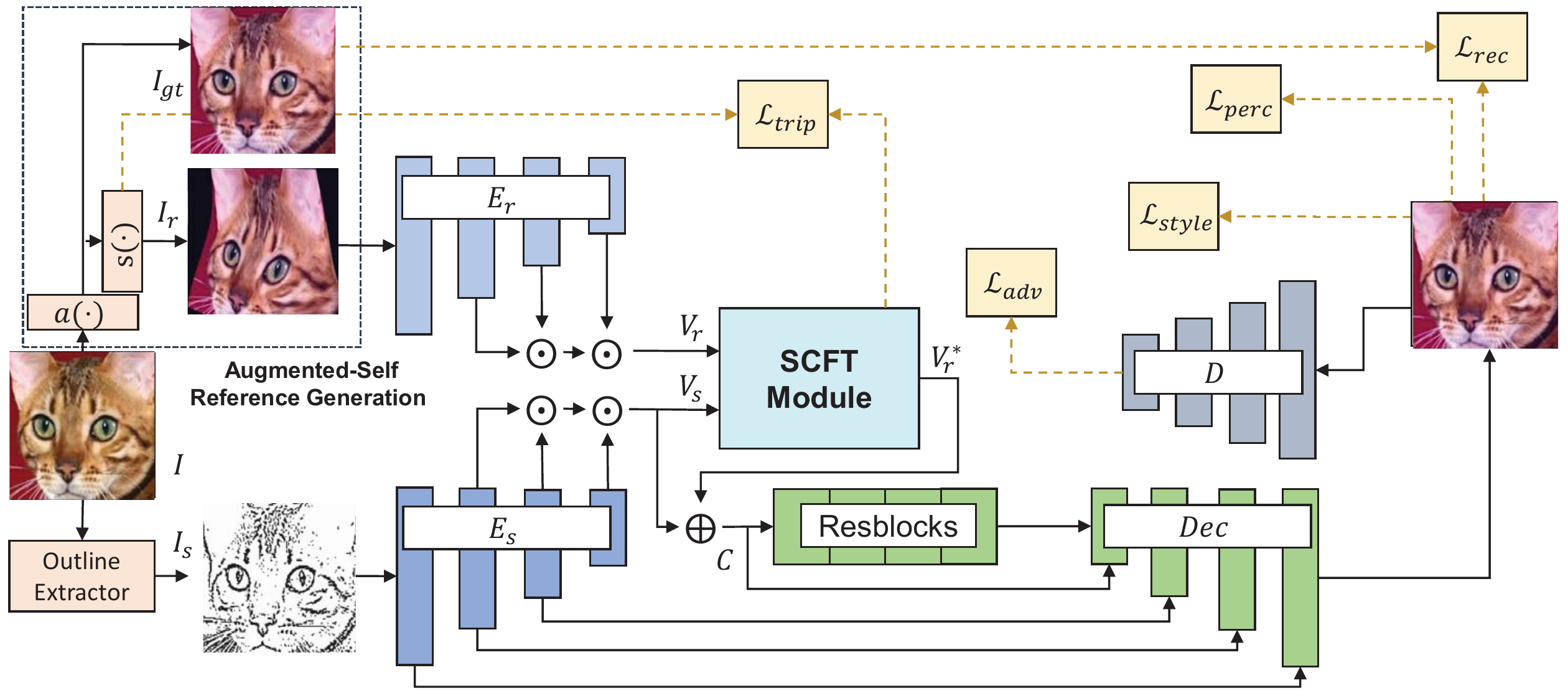}
\end{center}
   \caption{An overall workflow of our self-augmented learning process. }
\label{fig:main_overview}
\end{figure*}

\section{Proposed method}
 In this section, we present our proposed model in detail, as illustrated in Fig. \ref{fig:main_overview}. 
 We first describe overall workflow of the model and its two novel components called (1) Augmented-Self Reference Generation (Section~\ref{sec:augmented}) and (2) Spatially Corresponding Feature Transfer Module (Section~\ref{sec:spatial_attention_unit}). 
 We then present our loss functions in detail.

\subsection{Overall Workflow}
 As illustrated in Fig. \ref{fig:main_overview}, given a color image $I$ in our dataset, we first convert it into its sketch image $I_s$ using an outline extractor. 
 Additionally, we generate an augmented-self reference image $I_r$ by applying the thin plate splines (TPS) transformation.
 Taking these two images $I_s$ and $I_r$ as inputs, our model first encodes them into activation maps $f_s$ and $f_r$ using two independent encoders $E_s(I_s)$ and $E_r(I_r)$, respectively.
 
 % As our main approach to transfer the information from $I_r$ to $I_s$, we present a spatially corresponding feature transfer (SCFT) module inspired by a recently proposed self-attention mechanism~\cite{vaswani2017attention}, which computes dense correspondences between every pixel pair of $I_r$ to $I_s$. 
 To transfer the information from $I_r$ to $I_s$, we present a SCFT module inspired by a recently proposed self-attention mechanism~\cite{vaswani2017attention}, which computes dense correspondences between every pixel pair of $I_r$ to $I_s$. 
 Based on the visual mappings from SCFT, context features fusing the information between $I_r$ and $I_s$ passes through several residual blocks and our U-net-based decoder~\cite{ronneberger2015unet} sequentially to obtain the final colored output.

\subsection{Augmented-Self Reference Generation}
\label{sec:augmented}
 % To generate a semantically meaningful reference color image $I_r$ for a given sketch image $I_s$, we utilize its original color image $I$ itself by mainly applying to it two nontrivial transformations, appearance and spatial transformation. 
 To generate a reference color image $I_r$ for a given sketch image $I_s$, we apply to original color image $I$ two nontrivial transformations, appearance and spatial transformation. 
 Since $I_r$ is essentially generated from $I$, these processes guarantee that the useful information to colorize $I_s$ exists in $I_r$, which encourages the model to reflect $I_r$ in the colorization process. 
 The details on how these transformations operate are described as follows. 
 % Describe the appearance transformation:
 First, the appearance transformation $a(\cdot)$ adds a particular random noise per each of the RGB channel of $I$. 
 The resulting output $a(I)$ is then used as the ground truth $I_{gt}$ for the colorization output of our model. 
 The reason why we impose color perturbation for making reference is to prevent our model from memorizing color bias, which means that a particular object is highly correlated with the single ground truth color in train data (i,e., a red color for apples). 
 Given different reference in every iteration, our model should reconstruct different colored output for the same sketch, by leveraging 
 $I_r$ as the only path to restore $I_{gt}$.
 % Because the model should reconstruct different colored output for the same sketch image in every iteration, leveraging $I_r$ is the only path to restore $I_{gt}$.
 In other words, it encourages the model to actively utilize the information from $E_r$ not just from $E_s$ and generates reference-aware outputs at test time.
 % Describe the spatial transformation:
 Afterwards, we further apply the TPS transformation $s(\cdot)$, a non-linear spatial transformation operator to $a(I)$ (or $I_{gt}$), resulting in our final reference image $I_r$. 
 % In summary, the nontrivial transformations of $I$ into $I_r$ also prevent our model from lazily bringing the color in the same pixel position from $I_r$, while enforcing our model to identify semantically meaningful spatial correspondences even for a reference image with a spatially different layout, e.g., different poses.
 This prevents our model from lazily bringing the color in the same pixel position from $I_r$, while enforcing our model to identify semantically meaningful spatial correspondences even for a reference image with a spatially different layout, e.g., different poses.

\begin{figure}[h!]
\begin{center}
\includegraphics[width=\linewidth]{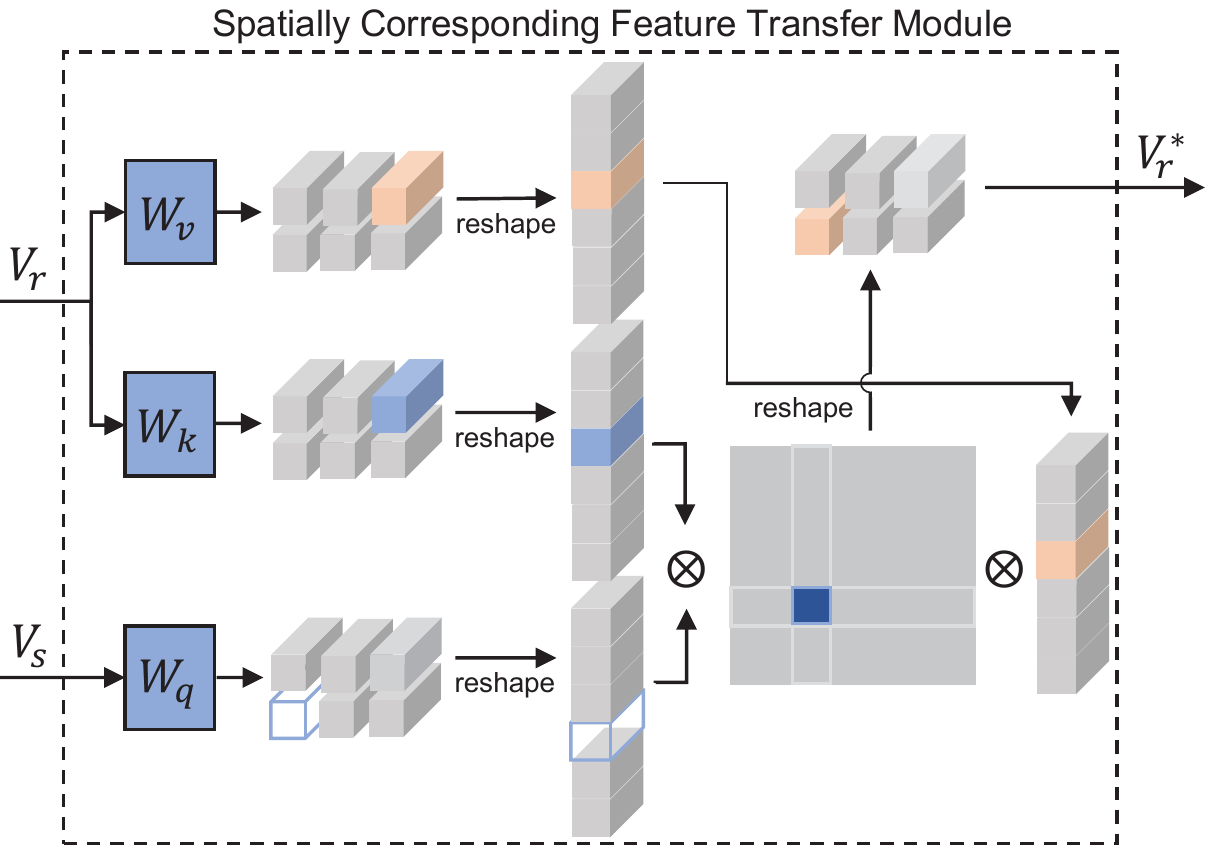}
\end{center}
   \caption{An illustration of spatially corresponding feature transfer (SCFT) module. SCFT establishes the dense correspondence mapping through attention mechanism.}
\label{fig:SCFT_view}
\end{figure}

\begin{figure*}[h!]
 \begin{center}
 \includegraphics[width=\linewidth]{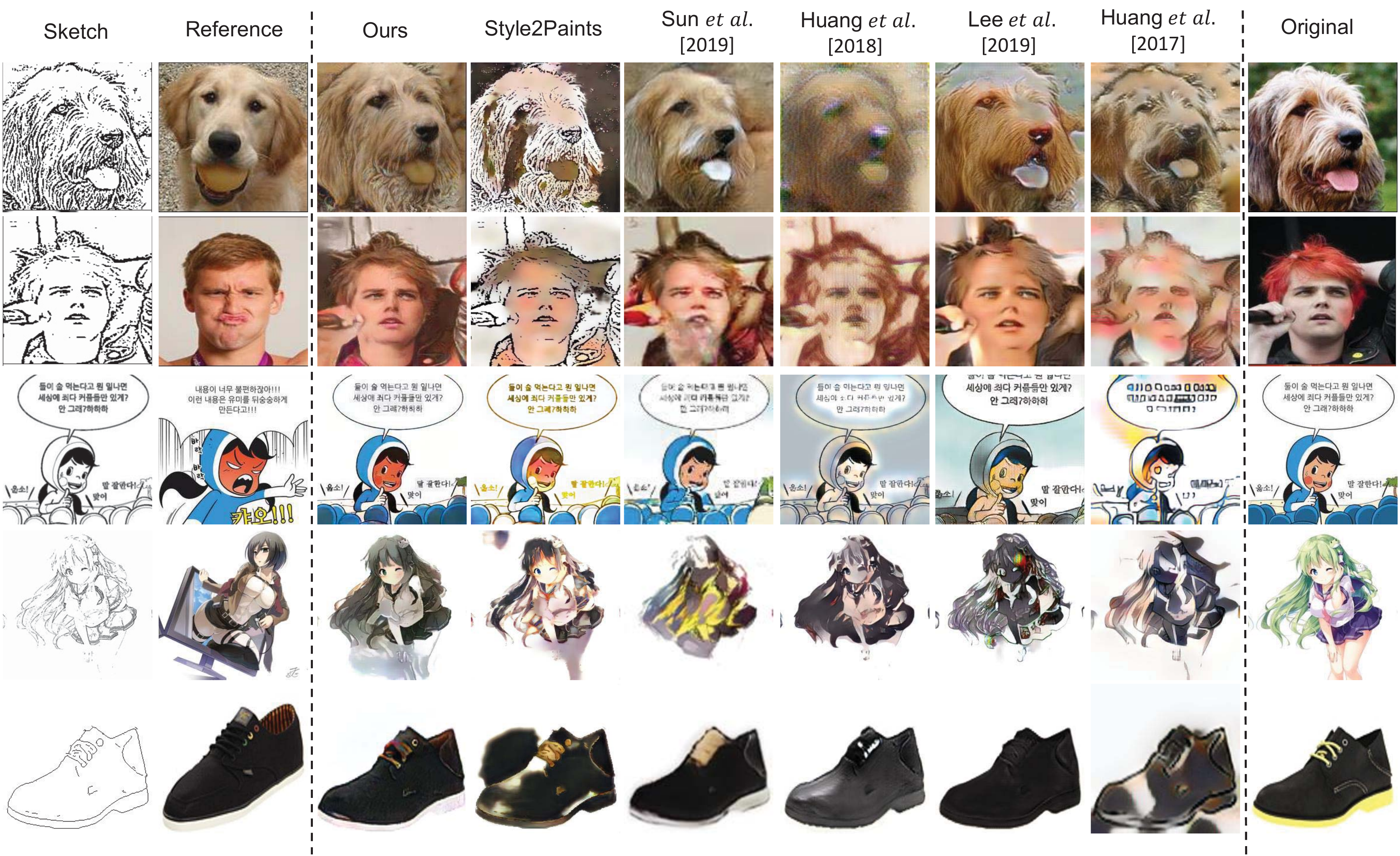}
 \end{center}
   \caption{Qualitative comparison of colorize results with the baselines trained on the wide range of datasets. Note that the goal of our task does not reconstruct the original image. All results are generated from the unseen images. Please refer to the supplementary material for details.}
 \label{fig:main_qualit}
 \end{figure*}

\subsection{Spatially Corresponding Feature Transfer}
\label{sec:spatial_attention_unit}
 The goal of this module is to learn (1) which part of a reference image to bring the information from as well as (2) which part of a sketch image to transfer such information to, i.e., transferring the information from where to where.
 Once obtaining this information as an attention map, our model transfers the feature information from a particular region of a reference to its semantically corresponding pixel of a sketch.
 
 To begin with, each of the two encoders $E_r$ and $E_s$ consists of $L$ convolutional layers, producing $L$ activation maps $(f^1, f^2, \cdots, f^L)$ including intermediate outputs.
 Now we downsample each of them to match the spatial size of $f^L$ and concatenate them along the channel dimensions, forming the final activation map $V$, i.e., 
 \begin{equation}
    V = \left[\varphi(f^1);\varphi(f^2);\cdots;f^{l_{p}} \right]
 \end{equation}
 where $\varphi$ denotes a spatially downsampling function of an input activation map $f^l \in \mathbb{R}^{h_l\times w_l\times c_l}$ to the size of $f^{l_p} \in \mathbb{R}^{h_p\times w_p\times c_p}$. $";"$ denotes the channel-wise concatenation operator.
 In this manner, we capture all the available low- to high-level features simultaneously. 
 
 Now we reshape $V$ into $\bar{V} = [v_1, v_2, \cdots, v_{hw}] \in \mathbb{R}^{d_{v}\times hw}$, where $v_i \in \mathbb{R}^{d_{v}}$ indicates a feature representation of the $i^{th}$ region of the given image and $d_{v}=\sum_{l=1}^L c_l$.
 We then obtain $v^s_{i}$ of $\bar{V_s}$ and $v^r_j$ of $\bar{V_r}$ from the outputs of the sketch encoder $E_s$ and the reference encoder $E_r$, respectively.
 Given $v^s_i$ and $v^r_j$, our model computes an attention matrix $\mathcal{A} \in \mathbb{R}^{hw\times hw}$ whose element $\alpha_{ij}$ is computed by the scaled dot product~\cite{vaswani2017attention}, followed by a softmax function within each row, i.e., 
 \begin{equation}
 \label{eq:attention_A}
    \alpha_{ij} = \underset{j}{\mbox{softmax}}\left( {(W_q v^s_i) \cdot (W_k v^r_j)\over \sqrt{d_v}}\right),
 \end{equation}
 where $W_q, W_k \in \mathbb{R}^{d_v \times d_v}$ represent the linear transformation matrix into a query and a key vector, respectively, in the context of a self-attention module, and $\sqrt{d_v}$ represents a scaling factor.
 $\alpha_{ij}$ is a coefficient representing how much information $v^s_i$ should bring from $v^r_j$.
 Now we can obtain the context vector $v^*_i$ of the position $i$ as 
 \begin{equation}
    v^*_i = \sum_{j}{\alpha_{ij}W_v v^{r_j}},
 \end{equation}
 where $W_v \in \mathbb{R}^{d_v\times d_v}$  is the linear transformation matrix into a value vector containing the color feature in a semantically related region of a reference image.
 
 Finally, $v^*_i$ is added to the original feature $v^s_i$ of a sketch image to form the feature vector enriched by the information of the corresponding region in the reference image, i.e., 
 \begin{equation}
    c_i = v^s_i + v^*_i
 \end{equation}
 $c_i$ is then fed into the decoder to synthesize a colored image.

%-------------------------------------------------------------------------
\begin{table*}[h!]
  \begin{center}
    \begin{tabular}{|l|| c | c | c | c | c | c | c  | c |}
    \hline
        ~ &\multicolumn{3}{c|}{ImageNet} & \multicolumn{1}{c|}{Human Face} & \multicolumn{2}{|c}{Comics} & \multicolumn{1}{|c|}{Hand-drawn} \\
        \hline
      Methods & Cat & Dog & Car &  CelebA & Tag2pix & Yumi's Cells & Edges$\rightarrow$Shoes \\
      \hline
      \hline
    %   Style2paints~\cite{style2paints} & 10.1 & b & 10.1 & 10.1 & 10.1  & 10.1 & 282.72 \\
      Sun \textit{et al.}~\cite{sun2019adversarial}& 160.65 & 168 & 192.00 & 75.66 & 122.14  & 72.45 & 124.98 \\
      Huang \textit{et al.}~\cite{huang2018multimodal} & 281.44 & 271.47 & 258.36 & 173.12  & 76.00 & 132.90 & 86.43 \\
      Lee \textit{et al.}~\cite{lee2019drit} & 151.52 & 172.22 & 70.07 & 68.43  & 91.65 & 63.34 & 109.29 \\
      Huang \textit{et al.}~\cite{huang2017arbitrary} & 257.39 & 268.69 & 165.84 & 160.22  & 97.40 & 148.52 & 190.16 \\
      % WCT & 10.1 & d & 10.1 & 10.1 & 10.1 & 10.1 & 10.1 & 10.1 \\
      \hline
      (a) Ours w/o $\mathcal{L}_{tr}$ & 77.39 & 109.49 & 54.07 & 53.58  & 47.68 & 51.34 & 79.85\\
      (b) Ours full & \textbf{74.12} & \textbf{102.83} & \textbf{52.23} & \textbf{47.15} & \textbf{45.34} & \textbf{49.29} & \textbf{78.32} \\
      \hline
    \end{tabular}

  \end{center}
  \caption{Quantitative comparisons over the datasets with existing baselines by measuring FID~\cite{heusel2017gans} score: a lower score is better.}
  \label{tab:quanti_baseline_main}
\end{table*}

\subsection{Objective Functions}
\label{sec:objectivefunctions}
 \noindent \textbf{Similarity-Based Triplet Loss.} When applying the spatial transformation $s(\cdot)$, each pixel value in the output image is represented as a weighted average of pixels in the input image, revealing the spatial correspondences of pixel pairs between $I_s$ and $I_r$. 
 In other words, we can obtain the full information of the weight $w_{ij}$, which represents how much the $i^{th}$ pixel position of the input image, or a query, is related to the $j^{th}$ pixel position of the output, or a key.
%  Then, the value of $\mbox{argmax}_i (w_{ij})$ can be considered as the pixel-to-pixel correspondence, which can work as the ground truth for supervising the most semantically related pixel position of the reference image to a particular pixel of an input image.
 Then, the value of $w_{ij}$ can be considered as the pixel-to-pixel correspondence, which can work as the groundtruth for supervising how semantically related the pixel of the reference to a particular pixel of sketch image.
 
 Utilizing this pixel-level correspondence information, we propose a similarity-based triplet loss, which is a variant of triplet loss~\cite{schroff2015triplet}, to directly supervise the affinity between the pixel-wise query and key vectors used to compute the attention map $\mathcal{A}$ in Eq.~\eqref{eq:attention_A}.
 %이부분은 supplementary로 뺌
 The proposed loss term is computed as
 \begin{equation}
    \mathcal{L}_{tr} = \mbox{max}(0, \left[ - S(v_q, v^p_k) + S(v_q, v^n_k) + \gamma \right]),
 \end{equation}
 where $S(\cdot, \cdot)$ computes the scaled dot product. Given a query vector $v_q$ as an anchor, $v^p_k$ indicates a feature vector sampled from the positive region, and $v^n_k$ is a negative sample. $\gamma$ denotes a margin, which is the minimum distance $S(v_q, v^p_k)$ and $S(v_q, v^n_k)$ should maintain.
 $\mathcal{L}_{tr}$ encourages the query representation to be close to the correct (positive) key representation, while penalizing to be far from the wrong (negatively sampled) one. 
 This loss plays a crucial role in directly enforcing our model to find the semantically matching pairs and reflect the reference color into the corresponding position. 
 
%  The reason we adopt triplet loss instead of commonly used losses such as $L_1$-loss is that the latter can overly penalize the affinities between semantically close but spatially distant query and key pixel pairs. For example, two distant query and key pixel representing \lq blue sky' have the same semantic meaning, which by $L_{1}$-loss can be penalized with supervision of no affinity score. This misleading result can be mitigated by only penalizing two cases: the semantically closest pair (positive sample) and randomly-sampled except it (negative sample), which is basically a triplet loss.
 The reason we adopt triplet loss instead of commonly used losses such as $L_1$-loss is that the latter can overly penalize the affinities between semantically close but spatially distant query and key pixel pairs.
 This misleading result can be mitigated by only penalizing two cases: the semantically closest pair (positive sample) and randomly-sampled except it (negative sample), which is basically a triplet loss.
 % Both quantitative and qualitative results are shown in Table~\ref{tab:quanti_baseline_main}, Table~\ref{tab:sc_psnr}, and Fig.~\ref{fig:ablation_objective}.
 
 We further conduct a user study to compare the effects of our triplet loss to another possible loss, i.e., $L_{1}$-loss and no supervision. Details about the experimental settings and results are explained in Section 6.2 in the supplementary material.
 
 \noindent \textbf{L1 Loss.} Since the groundtruth image $I_{gt}$ is generated as Section~\ref{sec:augmented}, we can directly impose a reconstruction loss to penalize the network for the color difference between the output and the ground truth image as below:
\begin{equation}
    \mathcal{L}_{rec} = \mathbb{E}\left[\parallel G(I_s, I_r) - I_{gt} \parallel_1 \right].
\end{equation}

 \noindent \textbf{Adversarial Loss.} The discriminator $D$, as an adversary of the generator, has an objective to distinguish the generated images from the real ones.
 The output of real/fake classifier $D(X)$ denotes the probability of an arbitrary image $X$ to be a real one.
 We adopt \textit{conditional GANs} which use both a generated sample and additional conditions~\cite{odena2017conditional, xu2018attngan, isola2017image}.
 % such as the class information~\cite{odena2017conditional}, text description~\cite{xu2018attngan}, and an input image~\cite{isola2017image}.
 In this work, we leverage the input image $I_s$ as a condition for the adversarial loss since it is important to preserve the content of $I_s$ as well as to generate a realistic fake image.
 The loss for optimizing $D$ is formulated as a standard cross-entropy loss as
 \begin{equation}
    \begin{split}
    \mathcal{L}_{adv} & = \mathbb{E}_{I_{gt}, I_s} \left[\log D(I_{gt}, I_s) \right] \\ 
    & + \mathbb{E}_{I_s, I_r} \left[\log (1 - D(G(I_s, I_r), I_s)) \right].
    \end{split}
 \end{equation}

 \noindent \textbf{Perceptual Loss.}
 As shown in previous work~\cite{nazeri2019edgeconnect}, perceptual loss~\cite{johnson2016perceptual} encourages a network to produce an output that is perceptually plausible. This loss penalizes the model to decrease the semantic gap, which means the difference of intermediate activation maps between the generated output $\hat{I}$ and the ground truth $I_{gt}$ from the ImageNet~\cite{russakovsky2015imagenet} pre-trained network. We employ a perceptual loss using multi-layer activation maps to reflect not only high-level semantics but also low-level styles as
 \begin{equation}
    \mathcal{L}_{perc} = \mathbb{E}\left[
    \sum_{l}{\parallel \phi_l(\hat{I})-\phi_l(I_{gt}) \parallel_{1,1}} \right],
 \end{equation}
 where $\phi_l$ represents the activation map of the $l$`th layer extracted at the \textit{relu}$l$\_1 from the VGG19 network.

 \noindent \textbf{Style Loss.} 
 Sajjadi \textit{et al.}~\cite{sajjadi2017enhancenet} has shown that the style loss which narrow the difference between the covariances of activation maps is helpful for addressing checkerboard artifacts. Given $\phi_l \in \mathbb{R}^{C_l\times H_l\times W_l}$, the style loss is computed as
 \begin{equation}
    \mathcal{L}_{style} = \mathbb{E}\left[ \parallel \mathcal{G}(\phi_l(\hat{I})) - \mathcal{G}(\phi_l(I_{gt})) \parallel_{1,1} \right],
 \end{equation}
 where $\mathcal{G}$ is a gram matrix.

 In summary, the overall loss function for the generator $G$ and discriminator $D$ is defined as
 \begin{equation}
    \begin{split}
        \min_{G}\max_{D}\mathcal{L}_{total} & = 
        \lambda_{tr}\mathcal{L}_{tr} + \lambda_{rec}\mathcal{L}_{rec} 
        + \lambda_{adv}\mathcal{L}_{adv} \\
        & +\lambda_{perc}\mathcal{L}_{perc} + \lambda_{style}\mathcal{L}_{style}.
    \end{split}
 \end{equation}
 
%-------------------------------------------------------------------------

 \subsection{Implementation Details}
 We implement our model with the size of input image fixed in 256$\times$256 on every datasets. 
 For training, we set the coefficients for each loss functions as follows: $\lambda_{adv}=1$, $\lambda_{rec}=30$, $\lambda_{tr}=1$, $\lambda_{perc}=0.01$, and $\lambda_{style}=50$. 
 We set the margin of the triplet loss $\gamma=12$ for overall data. 
 We use Adam solver~\cite{kingma2014adam} for optimization with $\beta_1=0.5$, $\beta_2=0.999$. 
 The learning rate of generator and discriminator are initially set to $0.0001$ and $0.0002$ for each. 
 The detailed network architectures are described in Section 6.5 of supplementary material.

%-------------------------------------------------------------------------
\section{Experiments}
 This section demonstrates the superiority of our approach on wide range of domain datasets (Section~\ref{sec:dataset}) including real photos, human face and anime (comics). %and Edges$\rightarrow$Shoes dataset. %well-known toy dataset.
 We newly present an evaluation metric, named SC-PSNR described in Section~\ref{sec:sp_psnr}, to measure the faithfulness of reflecting the style of the reference.
 Afterwards, we compare our method against the several baselines of related tasks quantitatively as well as qualitatively (Section~\ref{sec:comparison_baselines}).
 An in-depth analysis of our approach is described across Section~\ref{sec:analysis_loss_functions}-\ref{subsec:supp_visaulattentionmap}. 
 
 \subsection{Datasets}
 \label{sec:dataset}
 \noindent \textbf{Tag2pix Dataset.} We use Tag2pix dataset~\cite{kim2019tag2pix}, which contains large-scale anime illustrations filtered from Danbooru2017~\cite{danbooru2017}, to train our model for comic domain. Although there are various tag labels on this dataset, we only utilize images to train the model owing to our self-supervised training scheme. It consists of one character object with white background images. We partition into 54,317 images for train, 6036 images for test and then combine source-reference pairs by randomly sampled from the test set for evaluation.
 
 \noindent \textbf{Yumi Dataset.} Like Yoo \textit{et al.}~\cite{yoo2019coloring}, we collect images from the online cartoon named \textit{Yumi's Cells} for the outline colorization of the anime domain. The dataset contains %coherent 
 repeatedly emerging characters across 329 episodes. With this limited %data
 variety of characters, the network is required to find the correct character matching even if there is no explicit character supervisions. We randomly split into a train set of 7,014 images and test set of 380 images, and then manually construct source-reference pairs from the testset to evaluate the performance of the models.
 
 \begin{table}[h!]
  \begin{center}
    \begin{tabular}{|l || c | c | c |}
    \hline
      ~ &\multicolumn{3}{c|}{SC-PSNR (dB)} \\
      \hline
      Methods &  Cat & Dog & Car \\
      \hline
      \hline
      % Style2paints~\cite{style2paints} & 10.1 & b & 10.1 \\
      Sun \textit{et al.}~\cite{sun2019adversarial}& 9.65 & 11.19 & 9.42 \\ 
      Huang \textit{et al.}~\cite{huang2018multimodal} & 10.33 & 12.67 & 8.45 \\
      Lee \textit{et al.}~\cite{lee2019drit} & 11.54 & 12.08 &  9.94\\
      Huang \textit{et al.}~\cite{huang2017arbitrary} & 9.25 & 9.49 & 7.77  \\
      % WCT & 10.1 & d & 10.1 & 10.1 & 10.1 & 10.1 & 10.1 & 10.1 \\
      \hline
      (a) Ours w/o $\mathcal{L}_{tr}$ & 12.76 & 13.73 & 10.56 \\
      (b) Ours full  & \textbf{13.23} & \textbf{14.37} & \textbf{11.34} \\
      \hline
    \end{tabular}
  \end{center}
  \caption{Quantitative comparisons over the SPair-71k with existing baselines by measuring SC-PSNR (dB) score: a higher score is better.}
  \label{tab:sc_psnr}
\end{table}
 
 \noindent \textbf{SPair-71k Dataset.} SPair-71k dataset~\cite{min2019spair}, which is manually annotated for a semantic correspondence task, consists of total 70,958 pairs of images from PASCAL 3D+~\cite{xiang2014beyond} and PASCAL VOC 2012~\cite{everingham2015pascal}. We select two non-rigid categories (cat, dog) and one rigid category (car), of which we can gather sufficient data points from ImageNet~\cite{russakovsky2015imagenet}. 
 Note that this dataset is used to measure SC-PSNR (Section.~\ref{sec:sp_psnr}) score only for the evaluation purpose. 
 
 \noindent \textbf{ImageNet Dataset.} As above-mentioned, we collect
 subclasses that correspond to three categories (i.e., cat, dog, car) from ImageNet~\cite{russakovsky2015imagenet} dataset and use them for training data. Images in each class are randomly divided into two splits with an approximate ratio of 9:1 for training and validation.
 
 \noindent \textbf{Human Face Dataset.} Our method can be applied to colorize a sketch image of human face domain as well. To support this claim, we leverage CelebA~\cite{liu2015celeba} dataset, which have commonly been used for image-to-image translation or style transfer tasks. Training and validation sets are composed as the ImageNet dataset are.
 
 \begin{figure*}
    \begin{center}
    \includegraphics[width=\linewidth]{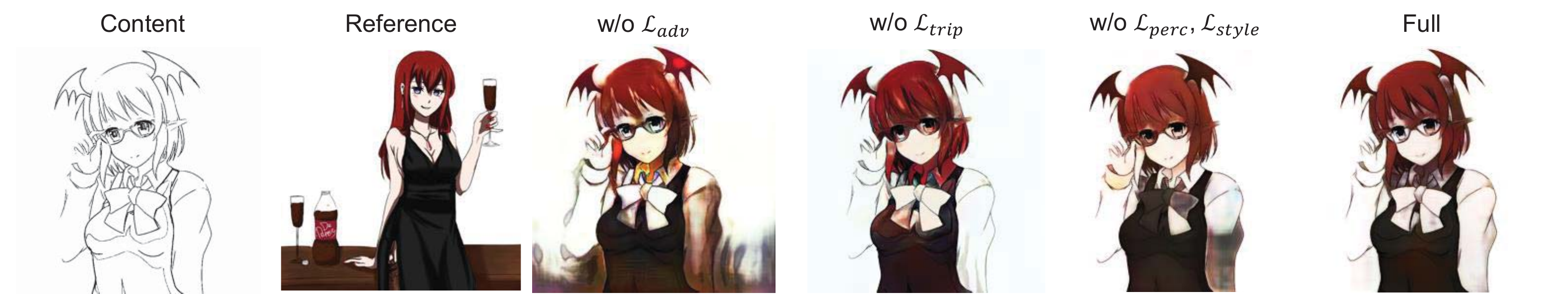}
    \end{center}
    \caption{A qualitative example presenting the effectiveness of different loss functions.}
    \label{fig:ablation_objective}
\end{figure*}

\begin{table*}[h!]
  \begin{center}
    \begin{tabular}{| l || c | c | c | c | c | c | c  | c }
    \hline
        ~ &\multicolumn{3}{c|}{ImageNet} & \multicolumn{1}{c|}{Human Face} & \multicolumn{2}{|c}{Comics} & \multicolumn{1}{|c|}{Hand-drawn} \\
        \hline
      Loss Functions & Cat & Dog & Car &  CelebA & Tag2pix & Yumi's Cells & Edges$\rightarrow$Shoes \\
      \hline
      \hline
    
      $\mathcal{L}_{rec}$& 82.10 & 143.76 & 68.45 & 77.70 & 58.00  & 52.86 & 91.10 \\
      $\mathcal{L}_{rec}+\mathcal{L}_{adv}$ & 78.56 & 110.86 & 56.54 & 54.75 & 48.71 & 51.96 & 82.55 \\
      $\mathcal{L}_{rec}+\mathcal{L}_{adv} + \mathcal{L}_{perc} + \mathcal{L}_{style}$ & 77.39 & 109.49 & 54.07 & 53.58  & 47.68 & 51.34 & 79.85 \\
      
      \hline
      $\mathcal{L}_{rec}+\mathcal{L}_{adv} + \mathcal{L}_{perc} + \mathcal{L}_{style} + \mathcal{L}_{tr}$& \textbf{74.12} & \textbf{102.83} & \textbf{52.23} & \textbf{47.15} & \textbf{45.34} & \textbf{49.29} & \textbf{78.32} \\
      \hline
    \end{tabular}

  \end{center}
  \caption{FID scores~\cite{heusel2017gans} according to the ablation of loss function terms described in Section~\ref{sec:analysis_loss_functions}. A lower score is better.}
  \label{tab:ablation_objective}
\end{table*}
 
 \noindent \textbf{Edges$\rightarrow$Shoes Dataset.} We use Edges$\rightarrow$Shoes dataset, which contains pairs of sketch-color shoes images that have been widely used in image-to-image translation tasks~\cite{lee2018diverse, huang2018multimodal} as well. This enables a valid evaluation between our method and existing unpaired image-to-image translation approaches.

 \subsection{Evaluation Metrics}
 \noindent \textbf{Semantically Corresponding PSNR.} \label{sec:sp_psnr}This work proposes a novel evaluation metric to measure how faithfully the model transfers the style of reference in the corresponding regions. 
 In the traditional automatic colorization setting where a groundtruth image is available, pixel-level evaluation metric, such as peak signal-to-noise ratio (PSNR), has been widely used.
 %its goal is to reconstruct an original image as a target label. 
 In reference-based colorization setting, however, there is no ground truth that have both the shape of the content and the style of the reference. 
 
 The key idea behind the semantically corresponding PSNR (SC-PSNR) is leveraging the datasets created for keypoint alignment tasks~\cite{everingham2015pascal, xiang2014beyond, min2019spair}, thereby providing patch-level groundtruth. 
 We use SPair-71k dataset~\cite{min2019spair} which contains semantically corresponding annotation pairs between two different images.
 Only the pixel values in a certain size of patch surrounding the corresponding keypoints of two images are used instead of the whole pixels for computing mean square error (MSE), and then PSNR is computed with the MSE.
 We refer to this measurement as the SC-PSNR.
%  The SC-PSNR is defined as
%  \begin{equation}
%     \begin{split}
%         \mbox{MSE} = \sum_{p}{(G(I_s, I_r)^{(p)}-I_r^{(p)})}\\
%         \mbox{SC-PSNR} = 10\log_{10}  \left( MAX^{2} \over MSE \right),
%     \end{split}
%  \end{equation}
% where ${p}$ denotes a ${p^{th}}$ patch around its key point and ${MAX}$ a maximum pixel value of the image.

%  Those datasets having semantically corresponding key points pairs. 
%  As all the positions and pixel values of matching pairs are given, we can now measure the PSNR on those regions
%  As we know each position of key point between two images, we can measure the patch-level distance of corresponding regions.
% Figure~\ref{fig:sc_psnr} shows perceptually plausible results queried by our SC-PSNR metric.

\begin{figure}[h!]
     \centering
     \includegraphics[width=\linewidth]{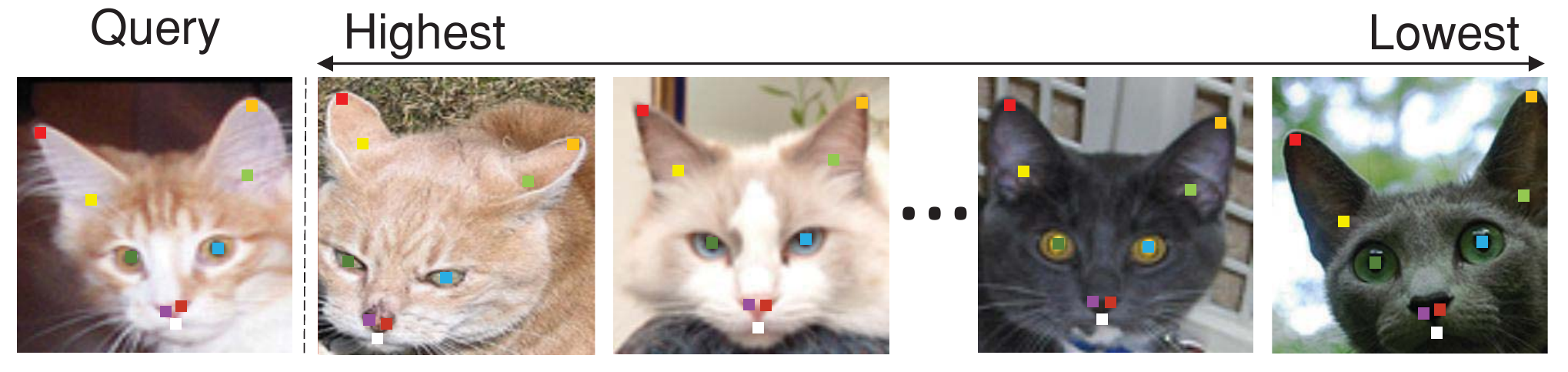}
     \caption{Different colors of points denote different keypoint annotations on cat face, e.g., eyes and noses.}
     \label{fig:sc_psnr}
 \end{figure}

 Fig.~\ref{fig:sc_psnr} shows first and last two examples of images queried by the leftmost image. The list of images are retrieved in a decreasing order of the value of SC-PSNR being computed with query. This figures demonstrates that this metric captures perceptually plausible distance of the pixel values between the keypoint regions of two images.  
 
 \noindent \textbf{\textit{Fr$\acute{e}$chet Inception Distance} (FID)~\cite{heusel2017gans}.} FID is a well-known metric for evaluating the performance of a generative model by measuring the Wasserstein-2 distance between the feature space representations of the real images and its generated outputs. A low score of FID indicates that the model generates the images with quality and diversity close to real data distribution.

%------------------------------------------------------------------------

\subsection{Comparisons to Baselines}
\label{sec:comparison_baselines}
 We compare our method against recent deep learning-based approaches on the various types of datasets both qualitatively and quantitatively.
 The baselines are selected from not only the colorization task~\cite{style2paints, sun2019adversarial} but also the related problems tackling multi-modal image generation, such as exemplar-guided image translation~\cite{huang2018multimodal, lee2019drit} and style transfer~\cite{huang2017arbitrary}.
 % The baselines include two colorization methods~\cite{style2paints, sun2019adversarial}, two image translation methods~\cite{huang2018multimodal, lee2019drit} that have tackled exemplar-guided image translation problem and one style transfer method~\cite{huang2017arbitrary} to compare the style transferability.
 
 Fig.~\ref{fig:main_qualit} shows the overall qualitative results of our model and other baselines on 5 different datasets.
 Datasets vary from real image domain like ImageNet or Human face dataset to sketch image domain like Edges$\rightarrow$Shoes, Yumi's Cells, and Tag2pix. The leftmost and second column are sketch and reference, respectively. 
 On every dataset our model brings the exact colors from the reference image and injects them into the corresponding position in the sketch. 
 For example, our model colorizes the character's face in third row with red color from the reference, while baselines tend not to fully transfer it. 
 Likewise, in fifth row, inner side of the shoes and shoe sole are elaborately filled with the color exactly referencing the exemplar image. 
 % In general, most of the models except ours perform well on at least one specific domain, however, our model has a superior quality of colorization on every domains.
 
 We report on Table~\ref{tab:quanti_baseline_main} the FID score calculated over the 7 different datasets. Our method outperforms the existing baselines by a large margin, demonstrating that our method has the robust capability of generating realistic and diverse images. Improved scores of our model with triplet loss indicates that $\mathcal{L}_{tr}$ plays a beneficial role in generating realistic images by directly supervising semantic correspondence.
 
 Table~\ref{tab:sc_psnr} presents the other quantitative comparisons in regard to the SC-PSNR scores as described in Section~\ref{sec:sp_psnr}.
 We measure SC-PSNR only over cat, dog and car dataset which are subclasses belonging to both ImageNet and SPair-71k~\cite{min2019spair}. Our method outperforms all the baseline models, demonstrating that our model is superior at establishing visual correspondences, and then generating suitable colors.
 
 We conduct a user study for human evaluation on our model and other existing baselines, as shown in Fig.~\ref{fig:user_study}. The detailed experimental setting is described in Section 6.2 in the supplementary material. Our model occupies a large percentage of Top1 and Top2 votes, indicating that our method better reflects the color from the reference and generates more realistic outputs than other baselines. 

 \subsection{Analysis of Loss Functions}
 \label{sec:analysis_loss_functions}
 We ablate the loss functions individually to analyze the effects of the functions qualitatively, as shown in Fig.~\ref{fig:ablation_objective} and quantitatively, as shown in Table ~\ref{tab:ablation_objective}. 
 When we remove $\mathcal{L}_{adv}$, output image contains inaccurate colors emerging in the background and dramatically appears unrealistic.
 Without $\mathcal{L}_{tr}$, character's back hair, forehead and ribbon tail are colorized with wrong color or even not colorized. The FID score in Table ~\ref{tab:ablation_objective} third row also represents that model generates unrealistic output. This degraded performance is due to the absence of supervision which encourages to match the semantically close regions between content and reference.
%  This is because $\mathcal{L}_{trip}$ encourages the model to distinguish the positive mapping from the irrelevant correspondence precisely.
 When we remove $\mathcal{L}_{perc}$ and $\mathcal{L}_{style}$, the colorization tends to produce color bleeding or visual artifacts since there is no constraint to penalize the model for the semantic difference between the model output and the ground truth.
 Image generated with full losses have exact colors in its corresponding regions with fewer artifacts.

 \subsection{Visualization of Attention Maps}
\label{subsec:supp_visaulattentionmap}
\begin{figure}[h]
\begin{center}
\includegraphics[width=1\linewidth]{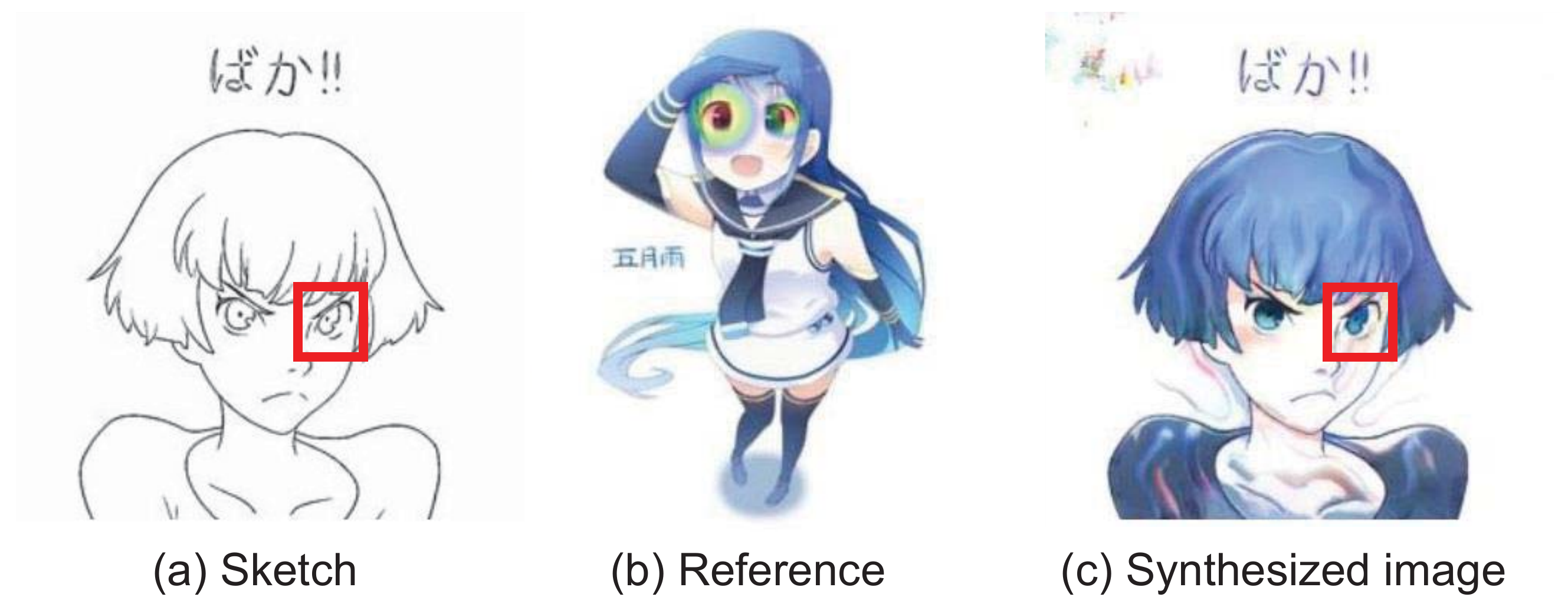}
\end{center}
   \caption{Visualization of our attention mechanism.}
\label{fig:supp_vis_attention}
\end{figure}
Fig.~\ref{fig:supp_vis_attention} shows an example of an attention map $\mathcal{A}$ learned by our SCFT module.  In this module, each pixel from the sketch is used as a query to retrieve the relevant local information from the reference. In the case of left-eye region as a query (red square in (a)), we visualize the top three, highly-attentive regions in the reference image  (a highlighted region in (b)). Based on this attention pattern, our model properly colorizes the left eye of a person in a sketch image (c) with blue color. For additional examples of visualizing attention maps for different sketch and reference images, we strongly encourage the readers to check out the Fig. 14 in the supplementary material for details. 

\begin{figure}[h]
\begin{center}
\includegraphics[width=1\linewidth]{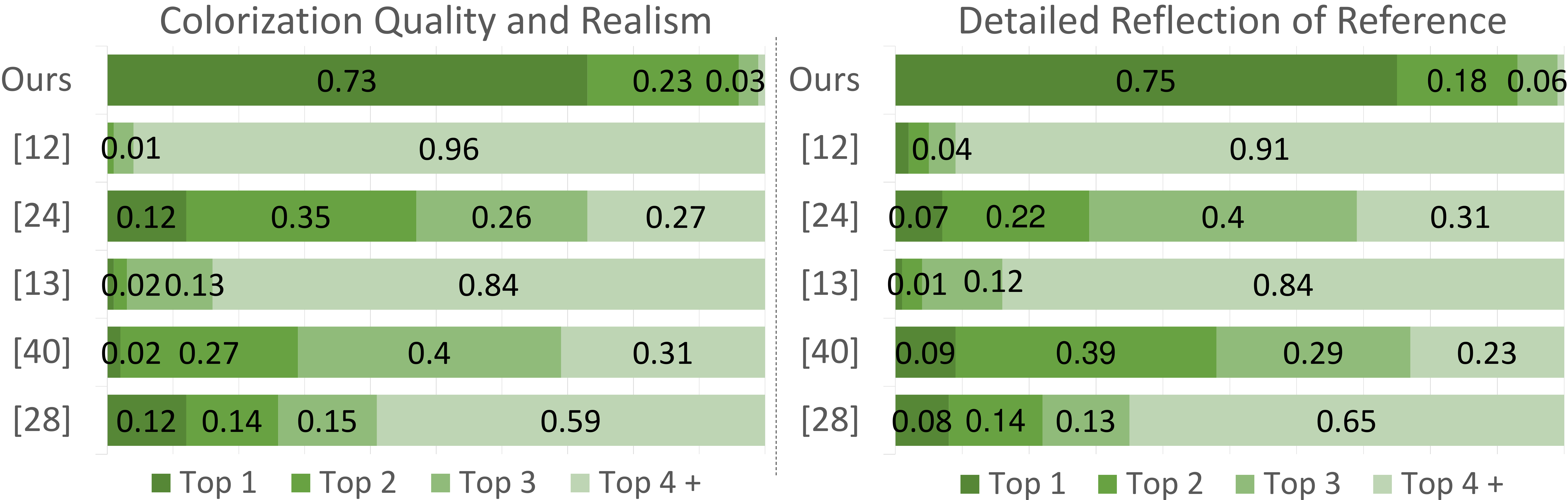}
\end{center}
   \caption{User study results. Percentage values are averaged over every datasets we experimented. Individual results are presented in Section 6.2 in supplementary material.}
\label{fig:user_study}
\end{figure}

%------------------------------------------------------------------------
\section{Conclusions}
This paper presents a novel training scheme, integrating  the augmented-self reference and the attention-based feature transfer module to directly learn the semantic correspondence for the reference-based sketch colorization task. 
Evaluation results demonstrate that our SCFT module exhibits the state-of-the-art performance over the diverse datasets, which demonstrates the significant potentials in practice. 
Finally, SC-PSNR, a proposed evaluation metric, effectively measures how the model faithfully reflects the style of the exemplar.
 
\vspace{2.0mm} 
 
\small{
\noindent \textbf{Acknowledgements.} 
This work was partially supported by Institute of Information \& communications Technology Planning \& Evaluation (IITP) grant funded by the Korea government(MSIT) (No.2019-0-00075, Artificial Intelligence Graduate School Program(KAIST)), the National Research Foundation of Korea (NRF) grant funded by the Korean government (MSIP) (No. NRF-2019R1A2C4070420), and the National Supercomputing Center with supercomputing resources including technical support (KSC-2019-CRE-0133). Finally, we thank all researchers at NAVER WEBTOON Corp. 
}

\clearpage

{\small
\bibliographystyle{latex/ieee_fullname}
\bibliography{main}

\begin{thebibliography}{10}\itemsep=-1pt

\bibitem{bugeau2013variational}
Aur{\'e}lie Bugeau, Vinh-Thong Ta, and Nicolas Papadakis.
\newblock Variational exemplar-based image colorization.
\newblock {\em IEEE Transactions on Image Processing}, 23(1):298--307, 2013.

\bibitem{charpiat2008automatic}
Guillaume Charpiat, Matthias Hofmann, and Bernhard Sch{\"o}lkopf.
\newblock Automatic image colorization via multimodal predictions.
\newblock In {\em ECCV}, pages 126--139, 2008.

\bibitem{chen2018sketchygan}
Wengling Chen and James Hays.
\newblock Sketchygan: Towards diverse and realistic sketch to image synthesis.
\newblock In {\em CVPR}, pages 9416--9425, 2018.

\bibitem{Chia:2011:SCI:2070781.2024190}
Alex Yong-Sang Chia, Shaojie Zhuo, Raj~Kumar Gupta, Yu-Wing Tai, Siu-Yeung Cho,
  Ping Tan, and Stephen Lin.
\newblock Semantic colorization with internet images.
\newblock {\em TOG}, 30(6):156:1--156:8, 2011.

\bibitem{geirhos2018imagenettrained}
Geirhos et al.
\newblock Imagenet-trained {CNN}s are biased towards texture; increasing shape
  bias improves accuracy and robustness.
\newblock {\em ICLR}, 2019.

\bibitem{everingham2015pascal}
Mark Everingham, SM~Ali Eslami, Luc Van~Gool, Christopher~KI Williams, John
  Winn, and Andrew Zisserman.
\newblock The pascal visual object classes challenge: A retrospective.
\newblock {\em International journal of computer vision}, 111(1):98--136, 2015.

\bibitem{gupta2012image}
Raj~Kumar Gupta, Alex Yong-Sang Chia, Deepu Rajan, Ee~Sin Ng, and Huang
  Zhiyong.
\newblock Image colorization using similar images.
\newblock In {\em MM}, pages 369--378, 2012.

\bibitem{danbooru2017}
Aaron~Gokaslan Gwern~Branwen.
\newblock Danbooru2017: A large-scale crowdsourced and tagged anime
  illustration dataset.
\newblock \url{https://www.gwern.net/Danbooru2017}, 2018.
\newblock [Online; accessed 22-03-2018].

\bibitem{he2016residual}
Kaiming He, Xiangyu Zhang, Shaoqing Ren, and Jian Sun.
\newblock Deep residual learning for image recognition.
\newblock In {\em CVPR}, pages 770--778, 2016.

\bibitem{he2018deep}
Mingming He, Dongdong Chen, Jing Liao, Pedro~V Sander, and Lu Yuan.
\newblock Deep exemplar-based colorization.
\newblock {\em TOG}, 37(4):47, 2018.

\bibitem{heusel2017gans}
Martin Heusel, Hubert Ramsauer, Thomas Unterthiner, Bernhard Nessler, and Sepp
  Hochreiter.
\newblock Gans trained by a two time-scale update rule converge to a local nash
  equilibrium.
\newblock In {\em NIPS}, pages 6626--6637, 2017.

\bibitem{huang2017arbitrary}
Xun Huang and Serge Belongie.
\newblock Arbitrary style transfer in real-time with adaptive instance
  normalization.
\newblock In {\em ICCV}, pages 1501--1510, 2017.

\bibitem{huang2018multimodal}
Xun Huang, Ming-Yu Liu, Serge Belongie, and Jan Kautz.
\newblock Multimodal unsupervised image-to-image translation.
\newblock In {\em ECCV}, pages 172--189, 2018.

\bibitem{ioffe2015batchnorm}
Sergey Ioffe and Christian Szegedy.
\newblock Batch normalization: Accelerating deep network training by reducing
  internal covariate shift.
\newblock In {\em ICML}, page 448–456, 2015.

\bibitem{isola2017image}
Phillip Isola, Jun-Yan Zhu, Tinghui Zhou, and Alexei~A Efros.
\newblock Image-to-image translation with conditional adversarial networks.
\newblock In {\em CVPR}, pages 1125--1134, 2017.

\bibitem{johnson2016perceptual}
Justin Johnson, Alexandre Alahi, and Li Fei-Fei.
\newblock Perceptual losses for real-time style transfer and super-resolution.
\newblock In {\em ECCV}, pages 694--711, 2016.

\bibitem{kekre2008color}
Hemant~B Kekre and Sudeep~D Thepade.
\newblock Color traits transfer to grayscale images.
\newblock In {\em 2008 First International Conference on Emerging Trends in
  Engineering and Technology}, pages 82--85, 2008.

\bibitem{kim2019tag2pix}
Hyunsu Kim, Ho~Young Jhoo, Eunhyeok Park, and Sungjoo Yoo.
\newblock Tag2pix: Line art colorization using text tag with secat and changing
  loss.
\newblock In {\em ICCV}, pages 9056--9065, 2019.

\bibitem{kingma2014adam}
Diederik~P Kingma and Jimmy Ba.
\newblock Adam: A method for stochastic optimization.
\newblock In {\em ICLR}, 2015.

\bibitem{yelamarthi2018zeroshot}
Sasi Kiran~Yelamarthi, Shiva Krishna~Reddy, Ashish Mishra, and Anurag Mittal.
\newblock A zero-shot framework for sketch based image retrieval.
\newblock In {\em ECCV}, 2018.

\bibitem{larsson2016learning}
Gustav Larsson, Michael Maire, and Gregory Shakhnarovich.
\newblock Learning representations for automatic colorization.
\newblock In {\em ECCV}, pages 577--593, 2016.

\bibitem{larsson2017colorization}
Gustav Larsson, Michael Maire, and Gregory Shakhnarovich.
\newblock Colorization as a proxy task for visual understanding.
\newblock In {\em CVPR}, pages 6874--6883, 2017.

\bibitem{lee2018diverse}
Hsin-Ying Lee, Hung-Yu Tseng, Jia-Bin Huang, Maneesh Singh, and Ming-Hsuan
  Yang.
\newblock Diverse image-to-image translation via disentangled representations.
\newblock In {\em ECCV}, pages 35--51, 2018.

\bibitem{lee2019drit}
Hsin-Ying Lee, Hung-Yu Tseng, Qi Mao, Jia-Bin Huang, Yu-Ding Lu, Maneesh Singh,
  and Ming-Hsuan Yang.
\newblock Drit++: Diverse image-to-image translation via disentangled
  representations.
\newblock {\em International Journal of Computer Vision}, 2020.

\bibitem{liu2019sketchgan}
Fang Liu, Xiaoming Deng, Yu-Kun Lai, Yong-Jin Liu, Cuixia Ma, and Hongan Wang.
\newblock Sketchgan: Joint sketch completion and recognition with generative
  adversarial network.
\newblock In {\em CVPR}, pages 5830--5839, 2019.

\bibitem{liu-2008-intrinsic}
Xiaopei Liu, Liang Wan, Yingge Qu, Tien-Tsin Wong, Stephen Lin, Chi-Sing Leung,
  and Pheng-Ann Heng.
\newblock Intrinsic colorization.
\newblock {\em TOG}, 27(5):152:1--152:9, 2008.

\bibitem{liu2015celeba}
Ziwei Liu, Ping Luo, Xiaogang Wang, and Xiaoou Tang.
\newblock Deep learning face attributes in the wild.
\newblock In {\em ICCV}, pages 3730--3738, 2015.

\bibitem{style2paints}
lllyasviel.
\newblock style2paints.
\newblock \url{https://github.com/lllyasviel/style2paints}, 2018.
\newblock [Online; accessed 22-03-2018].

\bibitem{lu2018image}
Yongyi Lu, Shangzhe Wu, Yu-Wing Tai, and Chi-Keung Tang.
\newblock Image generation from sketch constraint using contextual gan.
\newblock In {\em ECCV}, pages 205--220, 2018.

\bibitem{mao2017lsgan}
Xudong Mao, Qing Li, Haoran Xie, Raymond~YK Lau, Zhen Wang, and Stephen
  Paul~Smolley.
\newblock Least squares generative adversarial networks.
\newblock In {\em ICCV}, pages 2794--2802, 2017.

\bibitem{min2019spair}
Juhong Min, Jongmin Lee, Jean Ponce, and Minsu Cho.
\newblock Spair-71k: A large-scale benchmark for semantic correspondence.
\newblock {\em arXiv preprint arXiv:1908.10543}, 2019.

\bibitem{yumicells}
NaverWebtoon.
\newblock Yumi's cells.
\newblock \url{https://comic.naver.com/webtoon/list.nhn?titleId=651673}, 2019.
\newblock [Online; accessed 22-11-2019].

\bibitem{nazeri2019edgeconnect}
Kamyar Nazeri, Eric Ng, Tony Joseph, Faisal Qureshi, and Mehran Ebrahimi.
\newblock Edgeconnect: Structure guided image inpainting using edge prediction.
\newblock In {\em Proceedings of the IEEE International Conference on Computer
  Vision Workshops}, pages 0--0, 2019.

\bibitem{odena2017conditional}
Augustus Odena, Christopher Olah, and Jonathon Shlens.
\newblock Conditional image synthesis with auxiliary classifier gans.
\newblock In {\em ICML}, pages 2642--2651. JMLR. org, 2017.

\bibitem{ronneberger2015unet}
Olaf Ronneberger, Philipp Fischer, and Thomas Brox.
\newblock U-net: Convolutional networks for biomedical image segmentation.
\newblock In {\em International Conference on Medical image computing and
  computer-assisted intervention}, pages 234--241. Springer, 2015.

\bibitem{russakovsky2015imagenet}
Olga Russakovsky, Jia Deng, Hao Su, Jonathan Krause, Sanjeev Satheesh, Sean Ma,
  Zhiheng Huang, Andrej Karpathy, Aditya Khosla, Michael Bernstein, et~al.
\newblock Imagenet large scale visual recognition challenge.
\newblock {\em International journal of computer vision}, 115(3):211--252,
  2015.

\bibitem{sajjadi2017enhancenet}
Mehdi~SM Sajjadi, Bernhard Scholkopf, and Michael Hirsch.
\newblock Enhancenet: Single image super-resolution through automated texture
  synthesis.
\newblock In {\em ICCV}, pages 4491--4500, 2017.

\bibitem{sangkloy2017scribbler}
Patsorn Sangkloy, Jingwan Lu, Chen Fang, Fisher Yu, and James Hays.
\newblock Scribbler: Controlling deep image synthesis with sketch and color.
\newblock In {\em CVPR}, pages 5400--5409, 2017.

\bibitem{schroff2015triplet}
Florian Schroff, Dmitry Kalenichenko, and James Philbin.
\newblock Facenet: A unified embedding for face recognition and clustering.
\newblock In {\em CVPR}, pages 815--823, 2015.

\bibitem{sun2019adversarial}
Tsai-Ho Sun, Chien-Hsun Lai, Sai-Keung Wong, and Yu-Shuen Wang.
\newblock Adversarial colorization of icons based on contour and color
  conditions.
\newblock In {\em MM}, pages 683--691, 2019.

\bibitem{vaswani2017attention}
Ashish Vaswani, Noam Shazeer, Niki Parmar, Jakob Uszkoreit, Llion Jones,
  Aidan~N Gomez, {\L}ukasz Kaiser, and Illia Polosukhin.
\newblock Attention is all you need.
\newblock In {\em NIPS}, pages 5998--6008, 2017.

\bibitem{winnemoller2012xdog}
Holger Winnem{\"o}Ller, Jan~Eric Kyprianidis, and Sven~C Olsen.
\newblock Xdog: an extended difference-of-gaussians compendium including
  advanced image stylization.
\newblock {\em Computers \& Graphics}, 36(6):740--753, 2012.

\bibitem{xiang2014beyond}
Y. Xiang, R. Mottaghi, and S. Savarese.
\newblock Beyond pascal: A benchmark for 3d object detection in the wild.
\newblock In {\em 2014 IEEE Winter Conference on Applications of Computer
  Vision (WACV)}, pages 75--82, 2014.

\bibitem{xu2018attngan}
Tao Xu, Pengchuan Zhang, Qiuyuan Huang, Han Zhang, Zhe Gan, Xiaolei Huang, and
  Xiaodong He.
\newblock Attngan: Fine-grained text to image generation with attentional
  generative adversarial networks.
\newblock In {\em CVPR}, pages 1316--1324, 2018.

\bibitem{paintschainer}
Taizan Yonetsuji.
\newblock Paintschainer.
\newblock \url{https://paintschainer.preferred.tech/index_en.html}, 2017.
\newblock [Online; Accessed 22-03-2018].

\bibitem{yoo2019coloring}
Seungjoo Yoo, Hyojin Bahng, Sunghyo Chung, Junsoo Lee, Jaehyuk Chang, and
  Jaegul Choo.
\newblock Coloring with limited data: Few-shot colorization via memory
  augmented networks.
\newblock In {\em CVPR}, pages 11283--11292, 2019.

\bibitem{zhang2019deep}
Bo Zhang, Mingming He, Jing Liao, Pedro~V Sander, Lu Yuan, Amine Bermak, and
  Dong Chen.
\newblock Deep exemplar-based video colorization.
\newblock In {\em CVPR}, pages 8052--8061, 2019.

\bibitem{zhang2016colorful}
Richard Zhang, Phillip Isola, and Alexei~A Efros.
\newblock Colorful image colorization.
\newblock In {\em ECCV}, pages 649--666, 2016.

\bibitem{zhang2017real}
Richard Zhang, Jun-Yan Zhu, Phillip Isola, Xinyang Geng, Angela~S. Lin, Tianhe
  Yu, and Alexei~A. Efros.
\newblock Real-time user-guided image colorization with learned deep priors.
\newblock {\em TOG}, 36(4), 2017.

\end{thebibliography}
}

\clearpage

\def\thesection{\Alph{section}}

% \title{Supplementary Materials for ``Reference-Based Sketch Image Colorization using\\ Augmented-Self Reference and Dense Semantic Correspondence''}

\setcounter{section}{0}
%------------------------------------------------------------------------

\section{Supplementary Material}

This supplementary document presents additional details of the paper. % Section~\ref{subsec:supp_ablation_objectives} provides quantitative comparisons between the different combinations of objective functions. 
 Section~\ref{subsec:effectof_aggregation} discusses the effects of our spatially corresponding feature transfer mechanism with quantitative results.
 % Section~\ref{subsec:supp_visaulattentionmap} presents the visualization results of our attention mechanism, mainly shown in the attached video. 
 Section~\ref{subsec:supp_userstudy} demonstrates the human evaluation results that compare ours against baseline methods. 
 Afterwards, Section~\ref{subsec:supp_impledetails} reports implementation details including network architectures, the processes of generating augmented-self reference images, and other training details. 
 Comparisons to an existing study which shares similar network architectures are described in Section~\ref{subsec:supp_comparison_to_zhang}.
 Lastly, Section~\ref{subsec:supp_no_ref_case} addresses the case where a reference image does not exist. 
 Qualitative results generated by our method are also shown throughout the document.

% \begin{table*}[h!]
%   \begin{center}
%     \begin{tabular}{l|| c | c | c | c | c | c | c  | c }
%     \hline
%         ~ &\multicolumn{3}{c|}{ImageNet} & \multicolumn{1}{c|}{Human Face} & \multicolumn{2}{|c}{Comics} & \multicolumn{1}{|c|}{Hand-drawn} \\
%         \hline
%       Loss Functions & Cat & Dog & Car &  CelebA & Tag2pix & Yumi's Cells & Edge2Shoes \\
%       \hline
%       \hline
    
%       $\mathcal{L}_{rec}$& 82.10 & 143.76 & 68.45 & 77.70 & 58.00  & 52.86 & 91.10 \\
%       $\mathcal{L}_{rec}+\mathcal{L}_{adv}$ & 78.56 & 110.86 & 56.54 & 54.75 & 48.71 & 51.96 & 82.55 \\
%       $\mathcal{L}_{rec}+\mathcal{L}_{adv} + \mathcal{L}_{perc} + \mathcal{L}_{style}$ & 77.39 & 109.49 & 54.07 & 53.58  & 47.68 & 51.34 & 79.85 \\
      
%       \hline
%       $\mathcal{L}_{rec}+\mathcal{L}_{adv} + \mathcal{L}_{perc} + \mathcal{L}_{style} + \mathcal{L}_{trip}$& \textbf{74.12} & \textbf{102.83} & \textbf{52.23} & \textbf{47.15} & \textbf{45.34} & \textbf{49.29} & \textbf{78.32} \\
%       \hline
%     \end{tabular}

%   \end{center}
%   \caption{FID scores~\cite{heusel2017gans} due to the ablation of loss function terms described in Section 4.5 of the paper. A lower score is better.}
%   \label{tab:supp_quanti_objectives}
% \end{table*}

\begin{table*}[h!]
  \begin{center}
    \begin{tabular}{|l || c | c | c | c | c | c | c  | c }
    \hline
        ~ &\multicolumn{3}{c|}{ImageNet} & \multicolumn{1}{c|}{Human Face} & \multicolumn{2}{|c}{Comics} & \multicolumn{1}{|c|}{Hand-drawn} \\
        \hline
      Aggregation Method & Cat & Dog & Car &  CelebA & Tag2pix & Yumi's Cells & Edges2Shoes \\
      \hline
      \hline
      (a) Addition & 78.47 & 103.73 & 55.80 & 51.94 & 47.72 & 47.67 & 117.15 \\
      (b) AdaIN & 75.17 & 105.72 & 52.85 & 50.61 & 52.81 & \textbf{45.36} & 88.46 \\
      \hline
      (c) SCFT (ours) & \textbf{74.12} & \textbf{102.83} & \textbf{52.23} & \textbf{47.15} & \textbf{45.34} & 49.29 & \textbf{78.32} \\
      \hline
    \end{tabular}

  \end{center}
  \caption{FID scores~\cite{heusel2017gans} according to different aggregation methods.}
  \label{tab:supp_quanti_aggregation}
\end{table*}

% \subsection{Analysis of Spatially Corresponding Feature Transfer Module}
\subsection{Effects of Aggregation Methods}
\label{subsec:effectof_aggregation}
\begin{figure}[h]
\begin{center}
\includegraphics[width=\linewidth]{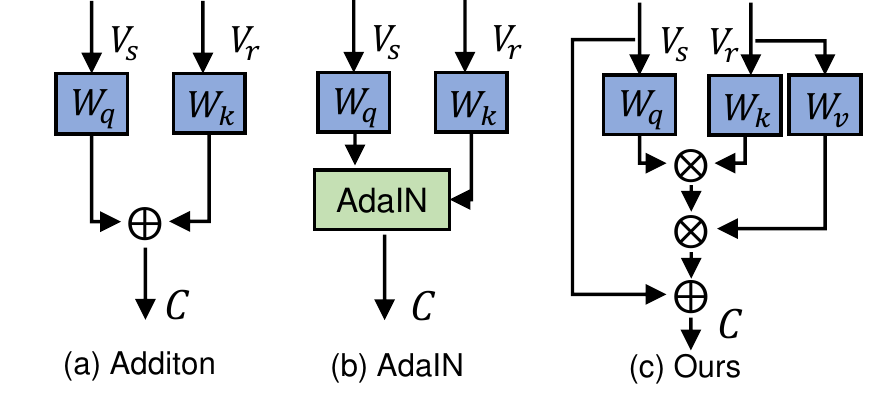}
\end{center}
\caption{Diagram of three types of aggregation methods. (a) Addition block, (b) AdaIN~\cite{huang2017arbitrary} block, (c) Ours (SCFT)}
\label{fig:ablation_threediagram}
\end{figure}

%  The key assumption behind SCFT is that adding information from the reference to the corresponding content features would be helpful to faithfully reflect the style of the exemplar in detail. 
 The key assumption behind SCFT is that integrating spatially aligned reference features with content features would help reflect the exact color from the reference into corresponding positions.  
 To prove this assumption, we compare our SCFT with two simple types of aggregation methods as shown in Fig.~\ref{fig:ablation_threediagram}. 
 Methods are as follows: (a) representations of the reference are simply added to the features of the content. 
 (b) AdaIN~\cite{huang2017arbitrary} is utilized to transfer the style of reference by aligning the channel-wise mean and variance of content to match those of reference. (c) our SCFT module.
 
\begin{figure}[h]
\begin{center}
\includegraphics[width=\linewidth]{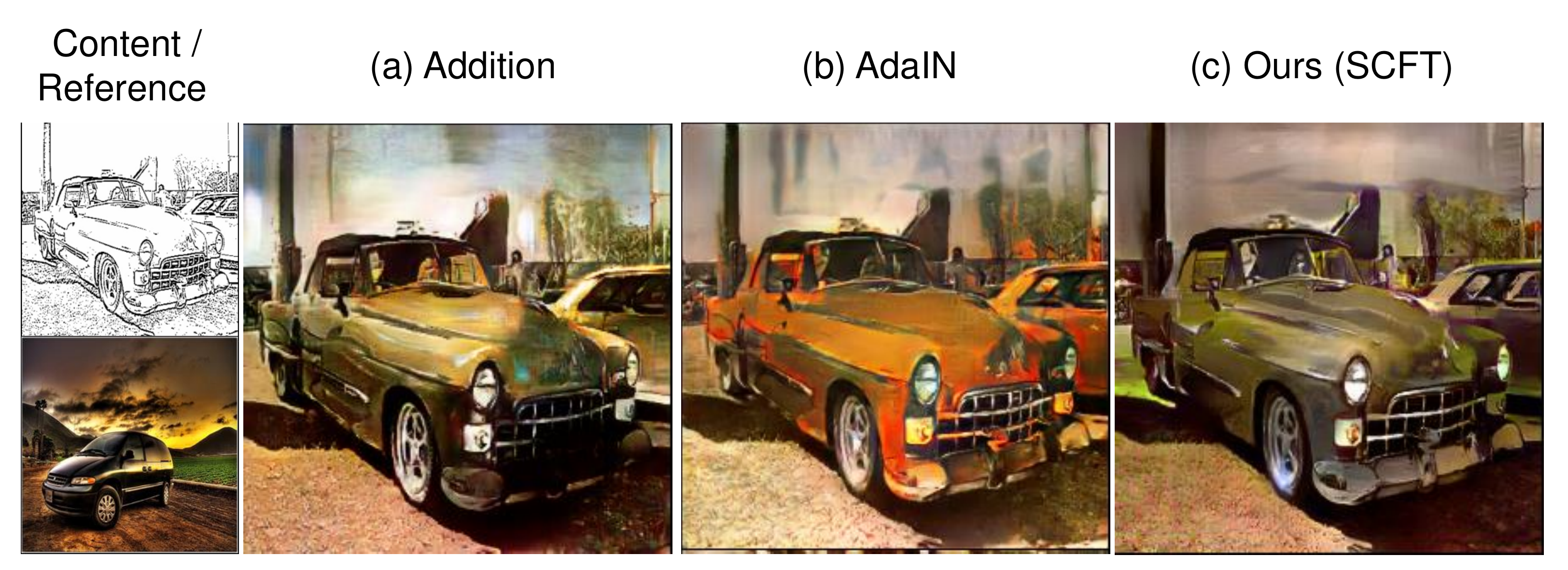}
\end{center}
\caption{A qualitative example obtained from three different aggregation methods as shown in Figure~\ref{fig:ablation_threediagram}.}
\label{fig:ablation_SCFT}
\end{figure}

 Qualitative comparison over three methods is shown in Fig.~\ref{fig:ablation_SCFT}. The leftmost column contains sketch and reference, while next three columns contain colorized images from (a), (b) and (c), respectively.
%  which is difficult to distinguish the black car from the sunset background. 
  Method (a) tends not to perfectly locate the corresponding regions and results in colorizing car with overly yellowish color, which is mainly background color in the exemplar. Method (b) totally ignores the spatially varying color information, thus colorizing with dominant color from the reference. (c) is superior to other methods in terms of color transferability to the corresponding position.
  
Quantitative results comparing these methods are represented in Table~\ref{tab:supp_quanti_aggregation}. The network with SCFT module produces the most realistic results over most of the datasets. This is because the SCFT module properly aligns the corresponding local regions between the sketch and the reference image by using the attention matrix 
$\mathcal{A}$. On the other hand, the method (a) and (b) are not capable of aligning the local features of the reference with those of the sketch, resulting in low FID scores.

In Yumi's Cells~\cite{yumicells} dataset, however, the SCFT module produced worse FID score than the others. The potential reason we assume is that the sketch and the reference we randomly pair for the inference time often contain different types of objects, e.g., Yumi (a human) and cells (non-human), which may have negatively impacted the colorization output. 
% When this noisy pair is given, our SCFT module fails. Specifically, there is a chance that the sketch only contains the 'cell' characters while the exemplar has 'human' characters, and vice versa. 
% With this condition, our module finds it hard to catch the matching regions between the two, eventually degrading the colorization quality. However, in case of Addition or AdaIN, it spreads the reference color widely into sketch image, which consequently outperforms SCFT module in average. A qualitative example is shown in Figure~\ref{fig:supp_ablation_aggregation}

\subsection{User Study}
\label{subsec:supp_userstudy}
We conduct two different human evaluation on the colorization outputs over various datasets. 
First, we randomly select ten sets of images per dataset, which contain the generated images from our method and other baselines. 
Second, we also randomly select ten sets of images for every dataset, and those contain the images obtained from the model trained with triplet loss, $L_{1}$-loss and no supervision for correspondence, respectively.
For both cases, participants with no prior knowledge in this work are asked to rank them in terms of two types of questions sequentially as follows:
\begin{myindentpar}{0.3cm}
\vspace*{-0.2cm}
\noindent$\bullet$ \textbf{Overall Colorization Quality and Realism} \\
How natural does the colorized image look? This question requires users to evaluate the overall quality of the generated colorization given an input sketch. The generated image should be perceptually realistic without any artifacts or color bleeding across sketches.
%이미지가 얼마나 사실적으로 보이는가? 이 질문은 유저들이 생성된 이미지가 얼마나 perceptually 이미지의 도메인과 가깝도록 만들어졌는지를 물어본다.

$\bullet$ \textbf{Detailed Reflection of Reference} \\
How well is the colors of the reference image is reflected to a given sketch part by part? This question asks users to determine whether the particular color from a reference is injected into the corresponding regions in the sketch. For example, given an comics character image with green hair wearing a blue shirt as a reference, the generated output is expected to contain these colors at its corresponding hair and clothing part, respectively.
\end{myindentpar}

% \begin{table}[h!]
%   \begin{center}
%     \begin{tabular}{l | c | c }
%     \hline
%       Methods &  Visual Realism & Reflection Style \\
%       \hline
%       \hline
%       % Style2paints~\cite{style2paints} & 10.1 & b & 10.1 \\
%       Style2paints~\cite{style2paints} & ?? & ??\\
%       Sun \textit{et al.}~\cite{sun2019adversarial}& ?? & ??\\ 
%       Huang \textit{et al.}~\cite{huang2018multimodal} & ?? & ??\\
%       Lee \textit{et al.}~\cite{lee2019drit} & ?? & ??\\
%       Huang \textit{et al.}~\cite{huang2017arbitrary} & ?? & ??  \\
%       % WCT & 10.1 & d & 10.1 & 10.1 & 10.1 & 10.1 & 10.1 & 10.1 \\
%       \hline
%       Ours full  & \textbf{??} & \textbf{??} \\
%       \hline
%     \end{tabular}
%   \end{center}
%   \caption{A result of user study.}
%   \label{tab:user_study}
% \end{table}

% based on the feedback from 20 users who have no prior knowledge of this work. 
% As seen in Table~\ref{tab:user_study}, our approach is ??\% more likely to be chosen as the top-rank output than the baseline methods. More specifically, ~~

As seen in Fig.~\ref{fig:subb_userstudy_1} and~\ref{fig:subb_userstudy_2}, superior measures indicate that our approach generates both more realistic and more faithfully colorized image than other methods. For both question type 1 and 2, it can be observed that our approach achieves the rank 1 votings more than 50\% over all the dataset we adopt for user study. When asked the first question on Comics domain dataset including Tag2pix~\cite{kim2019tag2pix} and Yumi's Cell~\cite{yumicells}, Style2Paints~\cite{style2paints} perform realistic generation quality comparable to our method with a small gap in top 1 rate. This notable measure is obtained as Style2Paints~\cite{style2paints} is a adept baseline especially on comic domain. However, the difference in top 1 rate increases as the users are asked to choose based on faithful colorization performance. The results demonstrate that our model utilizes the right color from the reference, which results in both realistic and exquisitely colorized output. 

The results in Fig.~\ref{fig:subb_userstudy_ablation} demonstrates that the model trained with triplet loss obtains more realistic and faithfully colorized outputs than with $L_{1}$-loss or no loss. Furthermore, along with the explanation of similarity-based triplet loss in Section 3.4 of the paper, these results support that the supervision for semantic correspondence with the $L_{1}$-loss leads to the inferior colorization performance even compared to the model without any supervision.

\subsection{Implementation Details}
\label{subsec:supp_impledetails}
This section provides the implementation details of our model, complementary to Section 3.5 of the paper.

\noindent \textbf{Augmented-Self Reference Generation}
To automatically generate a sketch image from an original color image, We utilize a widely-used algorithm called XDoG~\cite{winnemoller2012xdog}. The outputs, however, often involves superfluous edges, so in order to suppress them, we apply Gaussian blurring ($\sigma=0.7$) to the original images before extracting sketches. The appearance transformation $a(\cdot)$ adds randomly sampled value from a uniform distribution on [-50, 50] to each of the RGB channels of the original image. 

\noindent \textbf{Encoder} Our generator $G$ contains two types of encoder, $E_s$ and $E_r$. Both of them share the same architecture shown in Table~\ref{tab:supp_encoder_arch}, except for the number of input channels of the first layer, where $E_s$ takes a single-channel, binarized sketch input while $E_r$ takes a three-channel, RGB reference image. We utilize the an average pooling function for downsampling $\varphi$ in Section 3.3 of the paper.

\begin{table}[h]
  \begin{center}
    \begin{tabular}{ c | c }
    \hline
      Layer &  Encoder \\
      \hline
      \hline
      L1 & Conv(I:$C$,O:16,K:3,P:1,S:1), Leaky ReLU:0.2  \\
      \hline
      L2 & Conv(I:16,O:16,K:3,P:1,S:1), Leaky ReLU:0.2 \\
      \hline
      L3 & Conv(I:16,O:32,K:3,P:1,S:2), Leaky ReLU:0.2 \\
      \hline
      L4 & Conv(I:32,O:32,K:3,P:1,S:1), Leaky ReLU:0.2 \\
      \hline
      L5 & Conv(I:32,O:64,K:3,P:1,S:2), Leaky ReLU:0.2 \\
      \hline
      L6 & Conv(I:64,O:64,K:3,P:1,S:1), Leaky ReLU:0.2 \\
      \hline
      L7 & Conv(I:64,O:128,K:3,P:1,S:2), Leaky ReLU:0.2 \\
      \hline
      L8 & Conv(I:128,O:128,K:3,P:1,S:1), Leaky ReLU:0.2 \\
      \hline
      L9 & Conv(I:128,O:256,K:3,P:1,S:2), Leaky ReLU:0.2 \\
      \hline
      L10 & Conv(I:256,O:256,K:3,P:1,S:1), Leaky ReLU:0.2 \\
      \hline
    \end{tabular}
  \end{center}
  \caption{The network architecture of Encoder $E$. Conv denotes a convolutional layer. I, O, K, P, and S denote the number of input channels, the number of output channels, a kernel size, a padding size, and a stride size, respectively. }
  \label{tab:supp_encoder_arch}
\end{table}

\noindent \textbf{Resblocks} We place four stacked residual blocks~\cite{he2016residual} with a kernel size of 3 and a stride of 1. Batch normalization~\cite{ioffe2015batchnorm} follows each convolutional block, and ReLU is used as the activation function.

% \textbf{Decoder} To synthesize the output image using the latent features obtained from resblocks, the decoder block is composed of a nearest upsample function and a convolutional layer using Leaky ReLU 0.2.

\noindent \textbf{Discriminator} We adopt our discriminator architecture as PatchGAN~\cite{isola2017image}. We utilize the LSGAN~\cite{mao2017lsgan} objective for the stable training. 

\noindent \textbf{Training Details} For all the experiments, 
our network is trained using Adam optimizer~\cite{kingma2014adam} with $\beta_1=0.5$ and $\beta_2=0.999$. We set an initial learning rate for the generator as 0.0001 and that for the discriminator as 0.0002. We train the model for the first 100 epochs using the same learning rate, and then we linearly decay it to zero until the 200 epochs. 
%The coefficients for each loss functions are described in Section 4.2 of the paper. 
We set the margin value $\gamma=12$ for our triplet loss (Eq. 5 in the paper). The batch size is set as 16. The parameters of all our models are initialized according to the normal distribution which has a mean as 0.0 and a standard deviation as 0.02. 

\noindent \textbf{Baselines} We exploit Sun~\cite{sun2019adversarial} and Style2Paints~\cite{style2paints} as the sketch image colorization methods, Huang [2018]~\cite{huang2018multimodal}, and Lee~\cite{lee2019drit} as the image translation methods and Huang [2017]~\cite{huang2017arbitrary} as the style transfer method as our baselines. For Style2Paints~\cite{style2paints}, we generate the images based on the publicly available Style2Paints V3 in a  similar manner to Tag2pix~\cite{kim2019tag2pix}. For the other methods, we utilize the officially available codes to colorize images after training them on our datasets. 

\begin{figure}[h]
\begin{center}
\includegraphics[width=\linewidth]{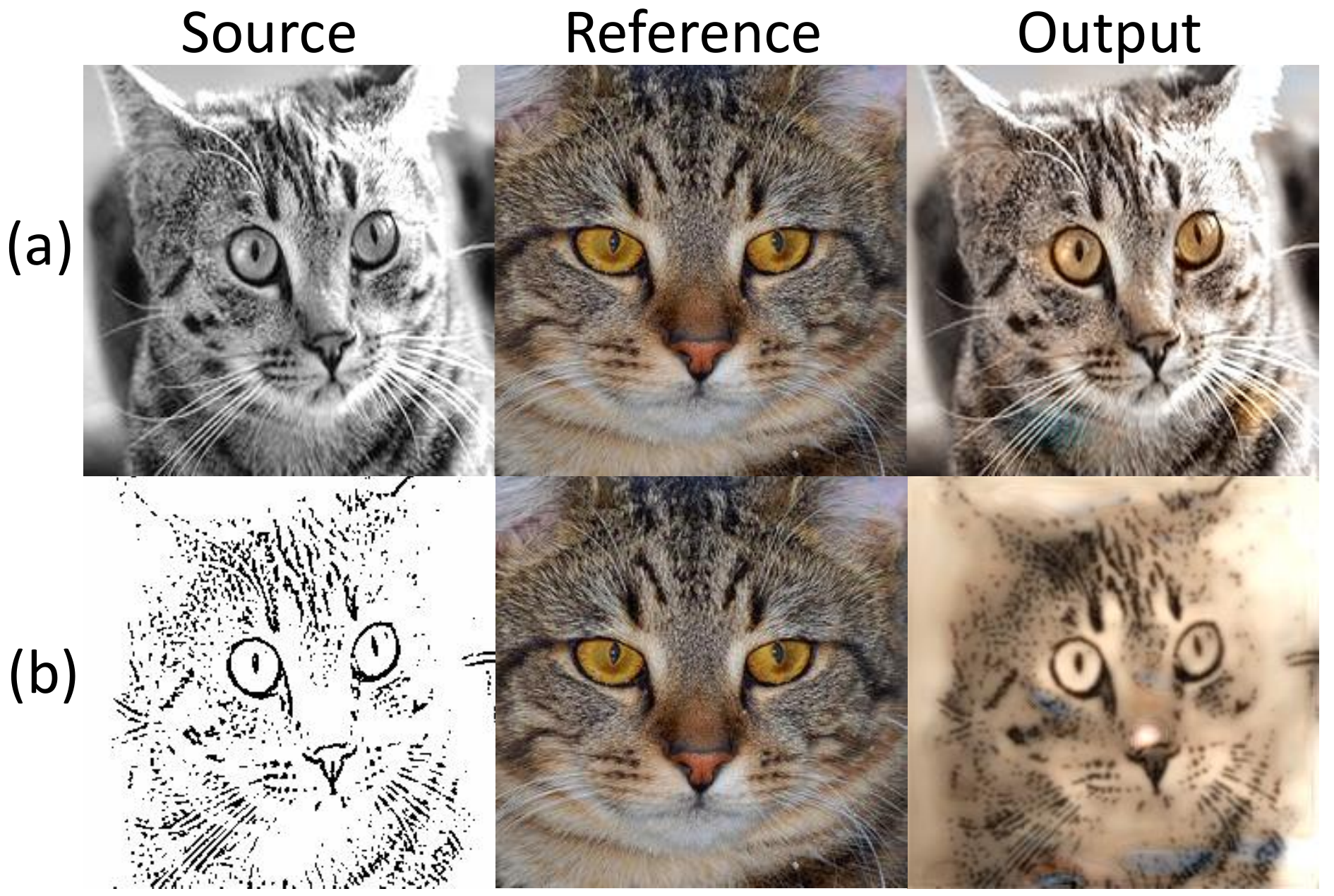}
\end{center}
   \caption{Qualitative results of our Zhang \textit{et al.}~\cite{zhang2019deep} given gray-scale source image (row (a)) and sketch image (row (b)). In contrast to the output in the row (a), output in (b) fails to colorize the eyes with the color from the reference and spreads the yellow color over the face.} 
\label{fig:supp_deep_video}
\end{figure}

\subsection{Comparison to Zhang \textit{et al.} (2019) ~\cite{zhang2019deep}.}
\label{subsec:supp_comparison_to_zhang}
 In this section, we discuss the detailed comparison between our method and Zhang \textit{et al.}. These two works have similarity in that they both exploit geometric distortion for data augmentation and semantic correspondence module for color guidance. However the significant difference of our model against Zhang \textit{et al.} lies in (1) direct supervision of semantic correspondence and (2) generalized attention module.
 
 \noindent \textbf{Direct supervision} Our model directly supervises the attention module via a triplet loss, which enables the optimization of the attention module in an end-to-end manner. This fully trainable encoder encourages to generate plausible results over a wide range of datasets from real-world photos to comic images, as show in Fig. 4 of the paper and Fig.~\ref{fig:supp_subqual_tag2pix}. In contrast, Zhang \textit{et al.} requires a pre-trained, already reliable attention module, which is only indirectly supervised via a so-called contextual loss. According to Geirhos \textit{et al.} (2019) ~\cite{geirhos2018imagenettrained}, the features extracted from the ImageNet pre-trained encoder may be severely degraded for a sketch image due to large domain shifts. In this sense, Zhang \textit{et al.}'s work may not be easily applicable to sketch image colorization tasks, and the examples of failure case are shown in Fig.~\ref{fig:supp_deep_video}. We reimplemented the code of Zhang \textit{et al.}, trained and tested the model over cat dataset. As this baseline exploits the ImageNet pre-trained encoder, row (a) shows that it produces the plausible colorized output given gray-scale source image. However, when given information scarce sketch image (row (b)), it fails to obtain the dense correspondence with the reference image, resulting in degraded output.
 
 \noindent \textbf{Generalized attention module} Inspired by the self-attention module in the Transformer networks, our attention module involves different query, key, and value mappings for flexibility, while Zhang \textit{et al.} use a relatively simple module.
 More importantly, in terms of value vectors, Zhang \textit{et al.} uses only raw color values, but ours uses all the available low- to high- level semantic information extracted from multiple layers.
 In this respect, ours is capable of transferring significantly richer contextual information than just low-level color information.

\begin{figure}[h]
\begin{center}
\includegraphics[width=\linewidth]{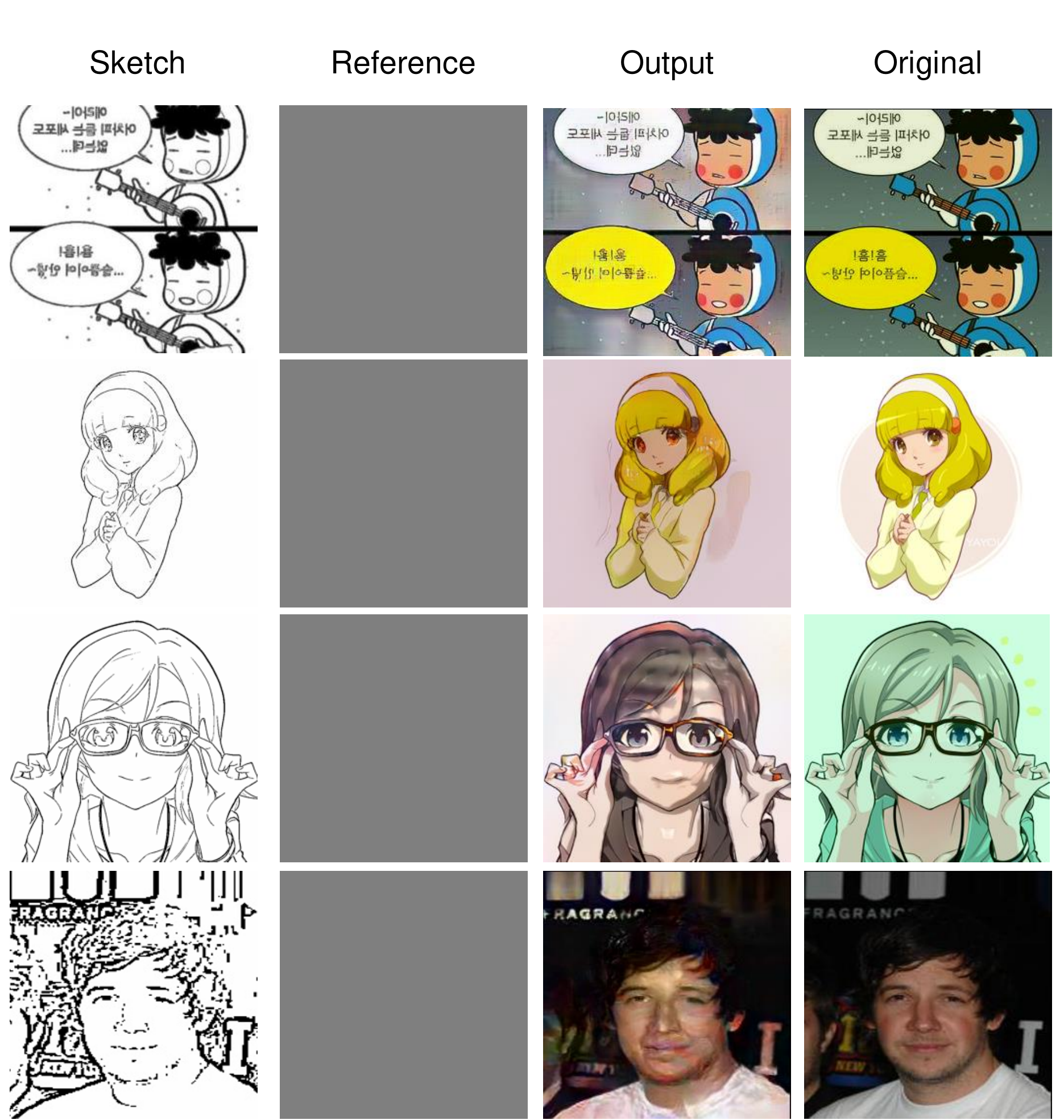}
\end{center}
\caption{A qualitative example when there is no reference image. Our model takes the first column image (sketch) as a target and the second column image (zero-filled reference) to synthesize the third column image (output). The results of first row, second-to-third rows, last row are obtained from our model trained for Yumi's Cells~\cite{yumicells}, Tag2pix~\cite{kim2019tag2pix}, and CelebA~\cite{liu2015celeba}, respectively.}
\label{fig:no_ref}
\end{figure}

\subsection{Colorization without reference.}
\label{subsec:supp_no_ref_case}
 Our main scope is focused on the colorization task with a reference available, but we can easily extend our method for no-reference cases by occasionally providing a zero-filled image as a reference to the networks during the training time. We feed the zero-filled image to our model as a reference with a ratio of 9:1 at the training time. 
 As shown in Fig.~\ref{fig:no_ref}, we confirm that  our network still generates a reasonable quality of colorization output at test time.
 In this case, the zero-filled reference image does not have any information to guide. 
 Therefore, the model is encouraged to synthesize an output image with colors that often appear in trainset conditioned on the sketch image.
 We recall that the main goal of this work is not restricted to generating the original image.
 
%  \twocolumn[{%
% \renewcommand\twocolumn[1][]{#1}%
% \maketitle
% \begin{center}
%     \centering
%     \includegraphics[width=\linewidth]{latex/images/supp_mainqual_celeba.pdf}
%     \captionof{figure}{Qualitative results of our method on the CelebA~\cite{liu2015celeba} dataset. Each row has the same content while each column has the same reference.}
%     % \caption{Qualitative results of our method on the CelebA dataset. }
%     \label{fig:our_celeba}
% \end{center}
% }]

\begin{figure*}[t]
\begin{center}
\includegraphics[width=\linewidth, height=0.57\linewidth]{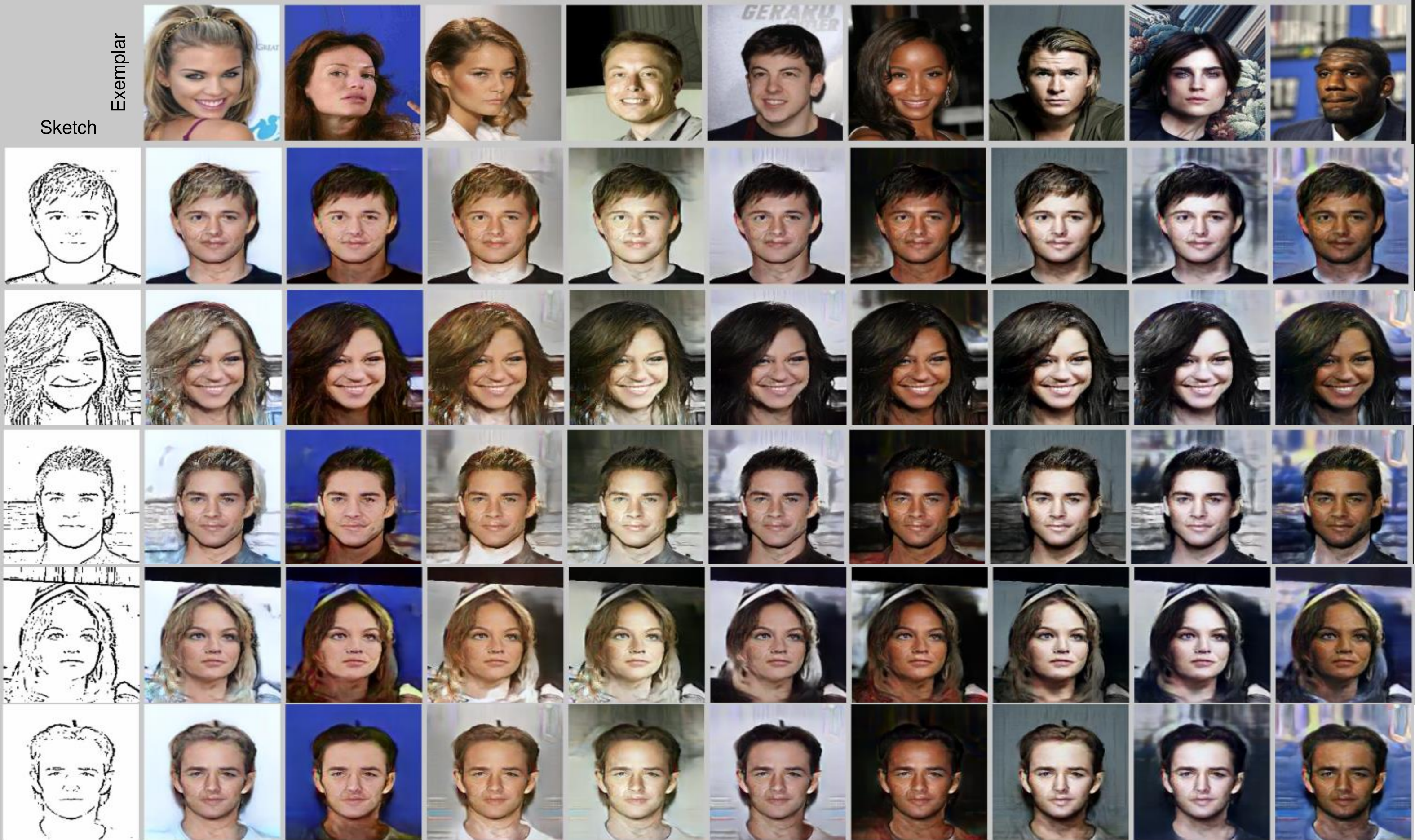}
\end{center}
   \caption{Qualitative results of our method on the CelebA~\cite{liu2015celeba} dataset.}
\label{fig:our_celeba}
\end{figure*}

\begin{figure*}[t]
\begin{center}
\includegraphics[width=\linewidth, height=0.57\linewidth]{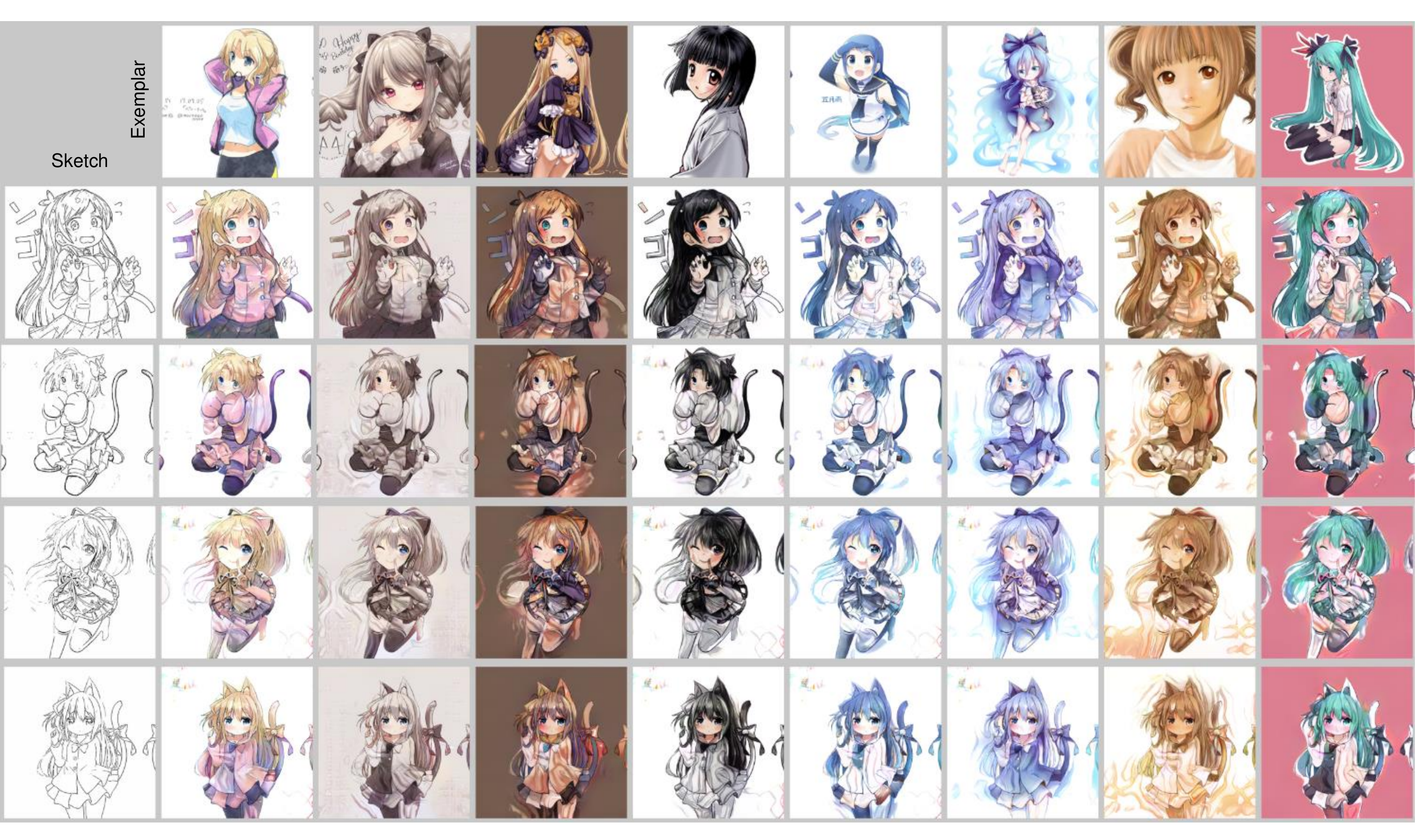}
\end{center}
   \caption{Qualitative results of our method on the Tag2pix~\cite{kim2019tag2pix} dataset.}
\label{fig:supp_mainqual_tag2pix}
\end{figure*}

\begin{figure*}
\begin{center}
\includegraphics[width=1\linewidth, height=0.57\linewidth]{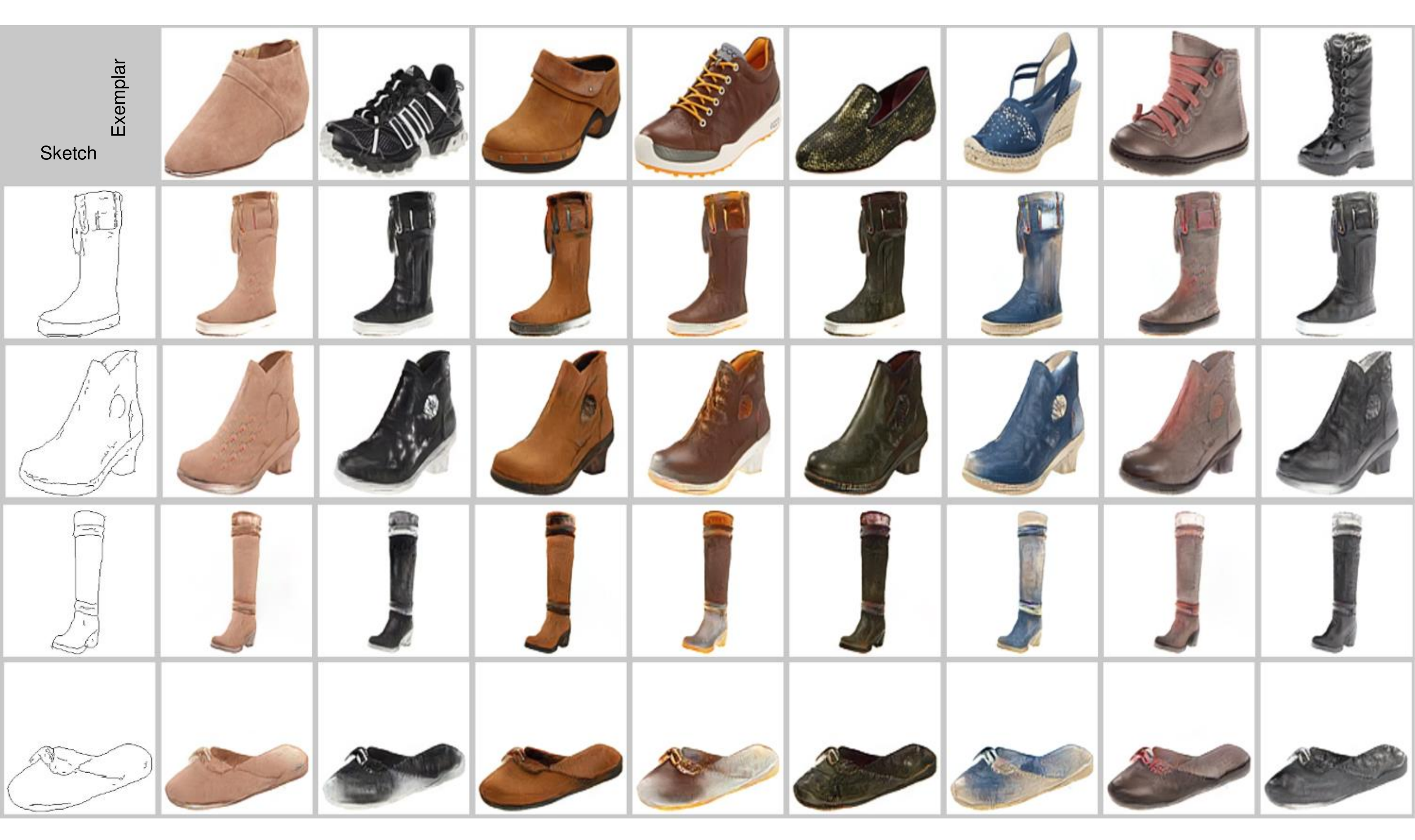}
\end{center}
   \caption{Qualitative results of our method on the Edges$\rightarrow$Shoes~\cite{isola2017image} dataset.}
\label{fig:supp_mainqual_edges2shoes}
\end{figure*}

\begin{figure*}
\begin{center}
\includegraphics[width=1\linewidth, height=0.57\linewidth]{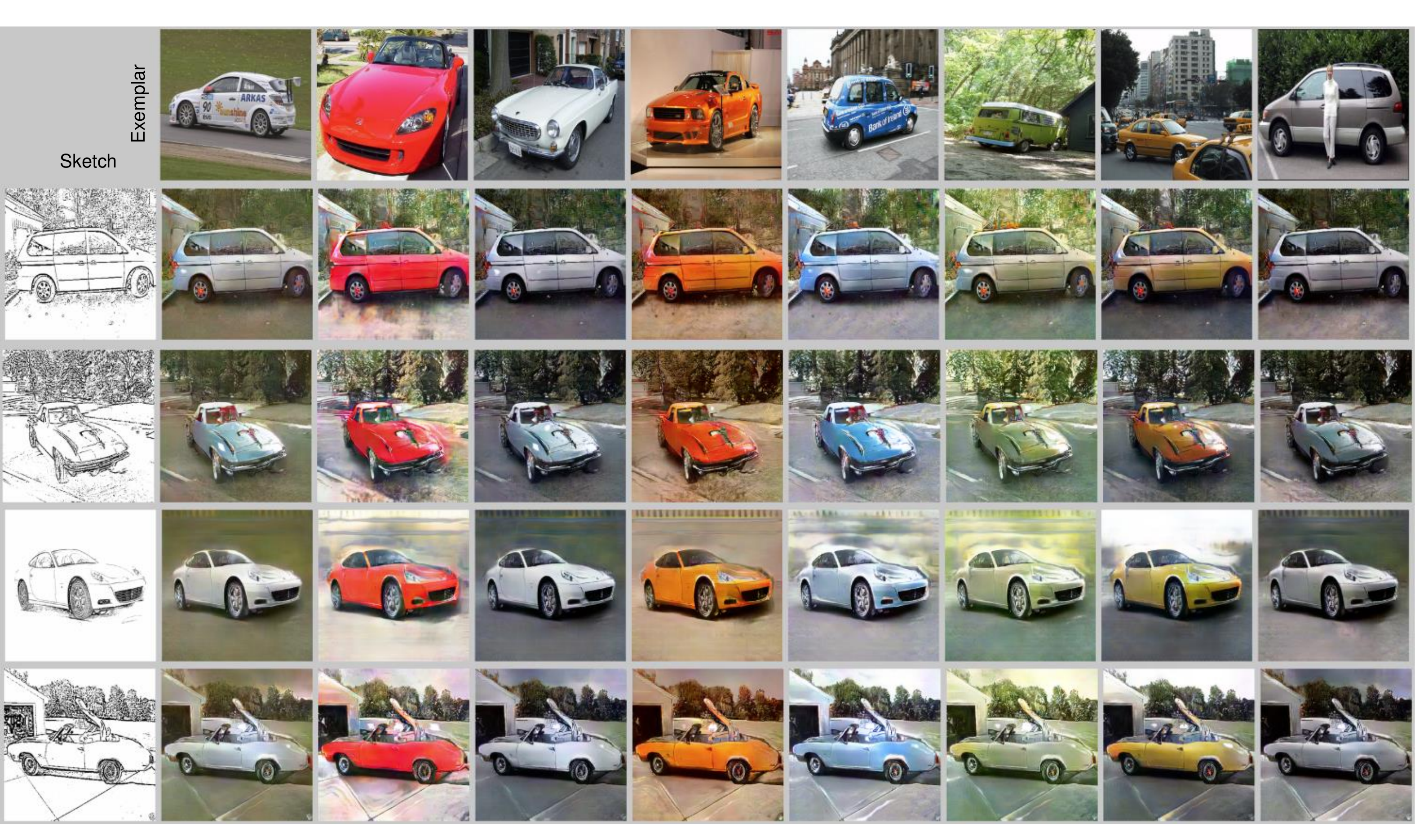}
\end{center}
   \caption{Qualitative results of our method on the ImageNet~\cite{russakovsky2015imagenet} dataset.}
\label{fig:supp_mainqual_car}
\end{figure*}

\begin{figure*}
\begin{center}
\includegraphics[width=1\linewidth, height=0.45\linewidth]{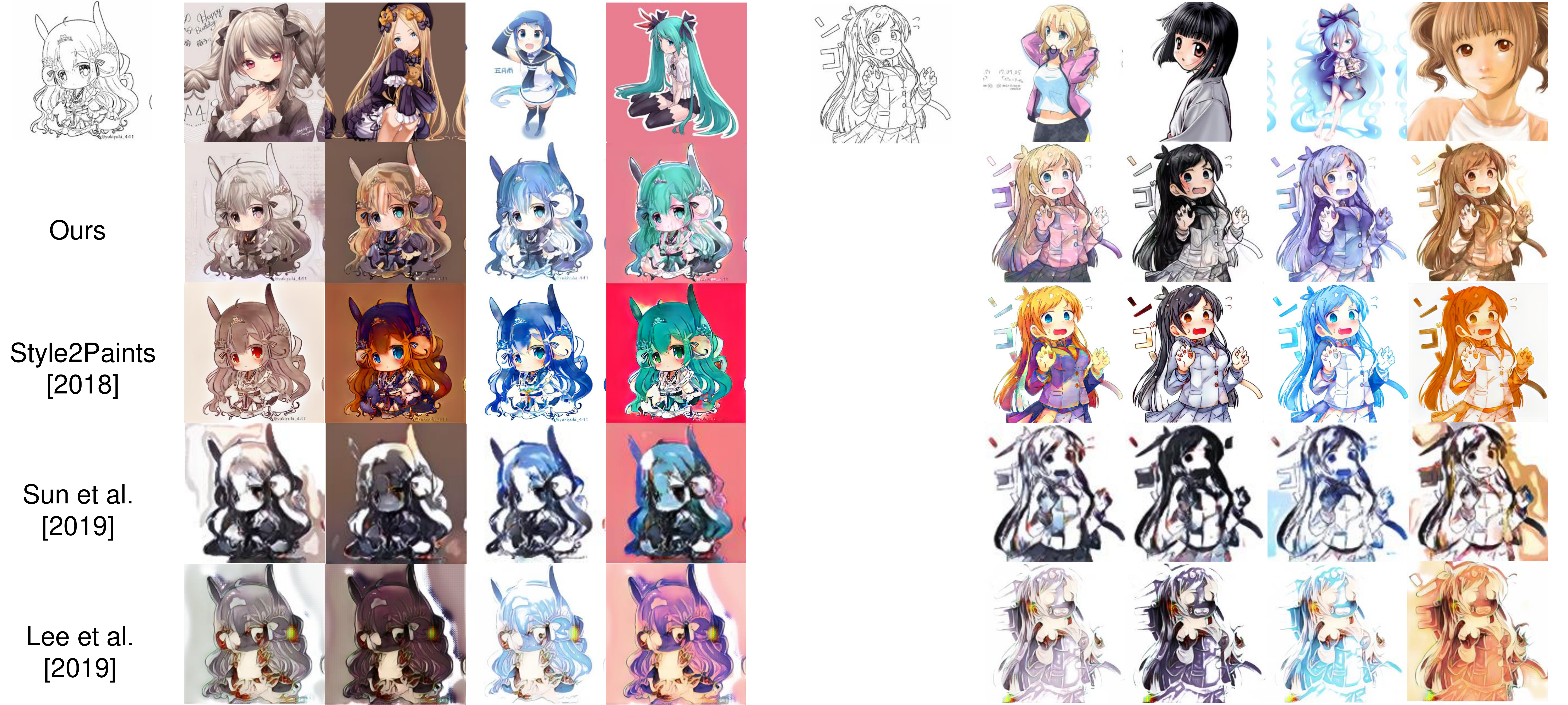}
\end{center}
   \caption{Qualitative comparisons with the baselines on the Tag2pix dataset.}
\label{fig:supp_subqual_tag2pix}
\end{figure*}

\begin{figure*}
\begin{center}
\includegraphics[width=1\linewidth, height=0.45\linewidth]{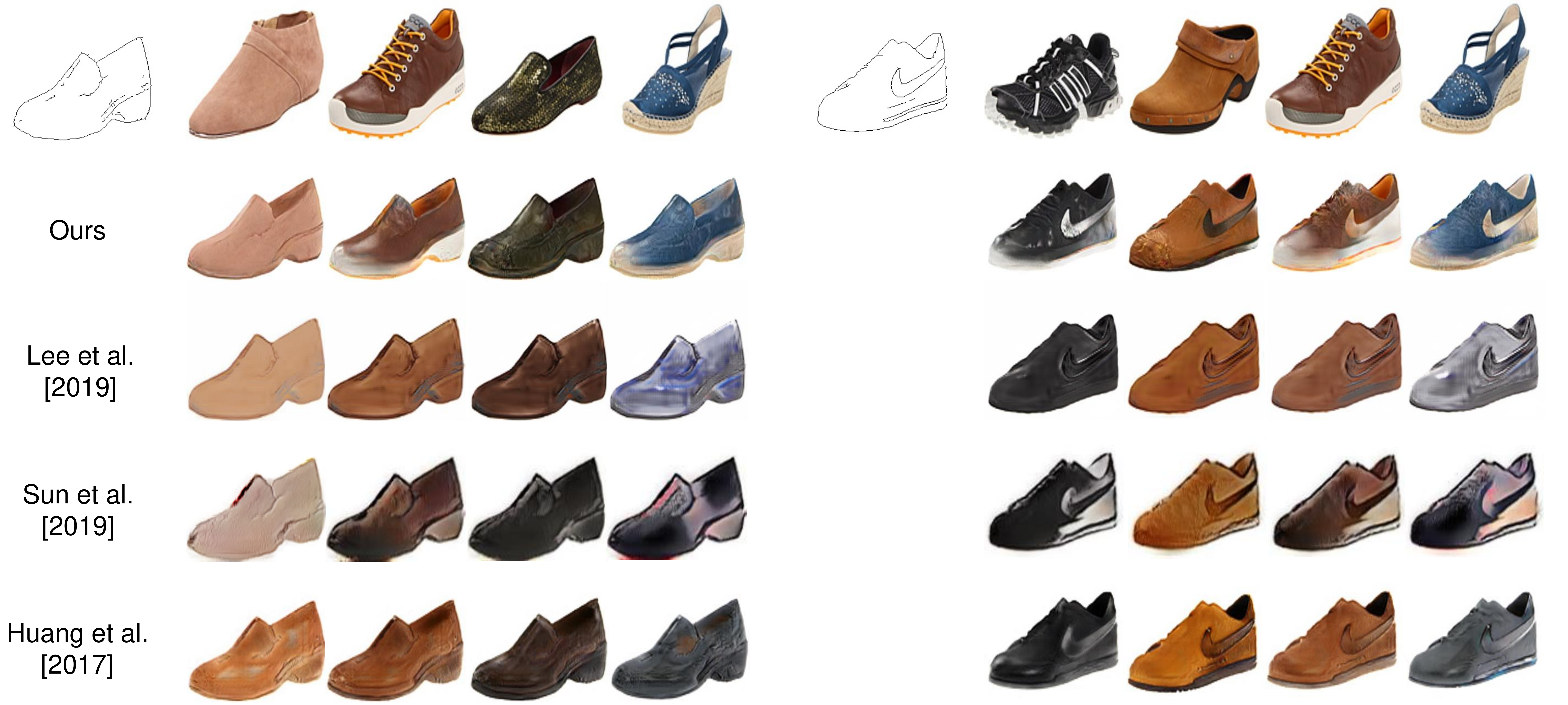}
\end{center}
   \caption{Qualitative results of our method on the Edges$\rightarrow$Shoes dataset.}
\label{fig:supp_subqual_e2s}
\end{figure*}

\begin{figure*}
\begin{center}
\includegraphics[width=0.85\linewidth]{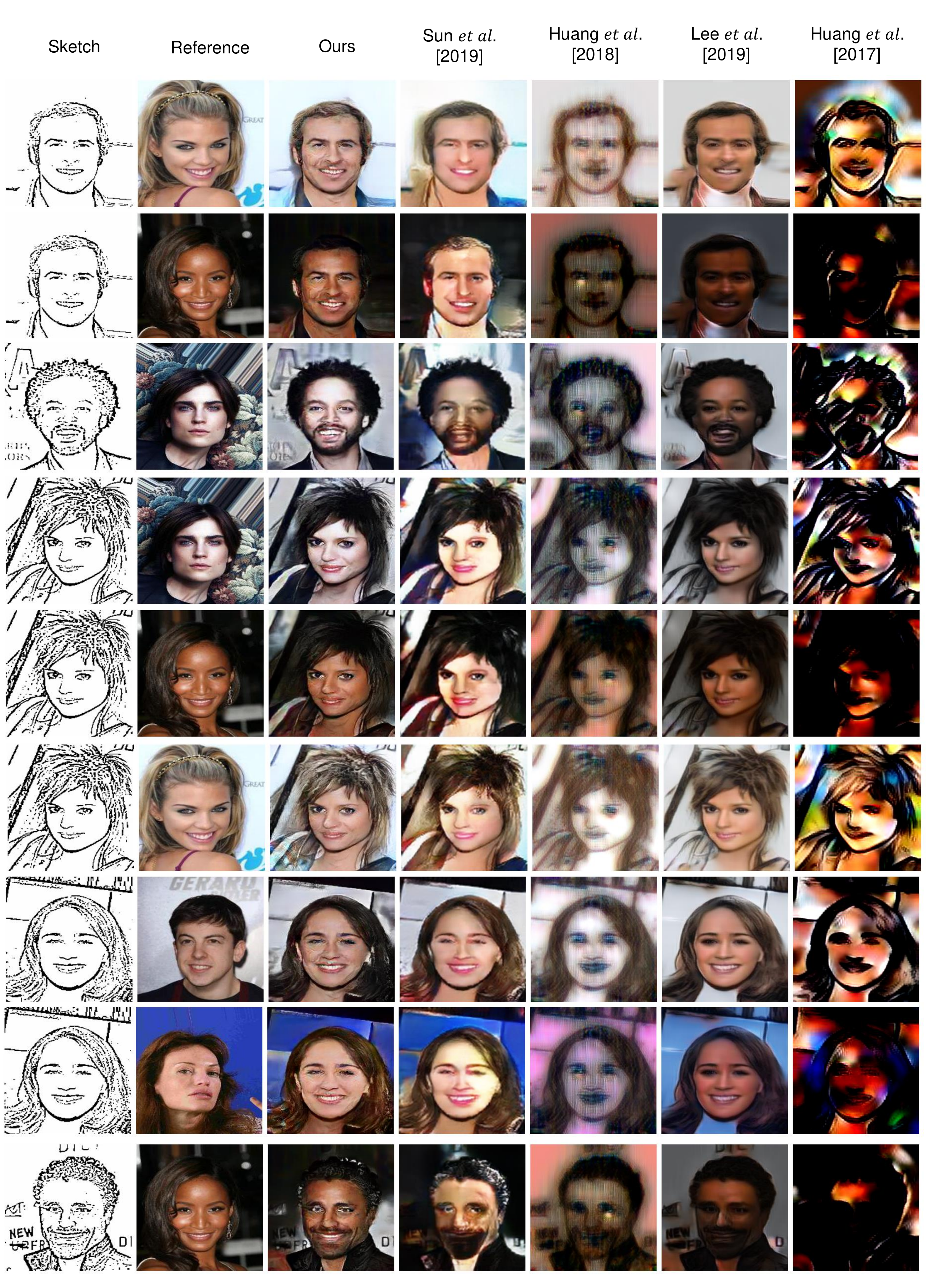}
\end{center}
   \caption{Qualitative comparisons with baselines on the CelebA dataset.}
\label{fig:supp_baselines_celeba}
\end{figure*}

\begin{figure*}
\begin{center}
\includegraphics[width=0.85\linewidth]{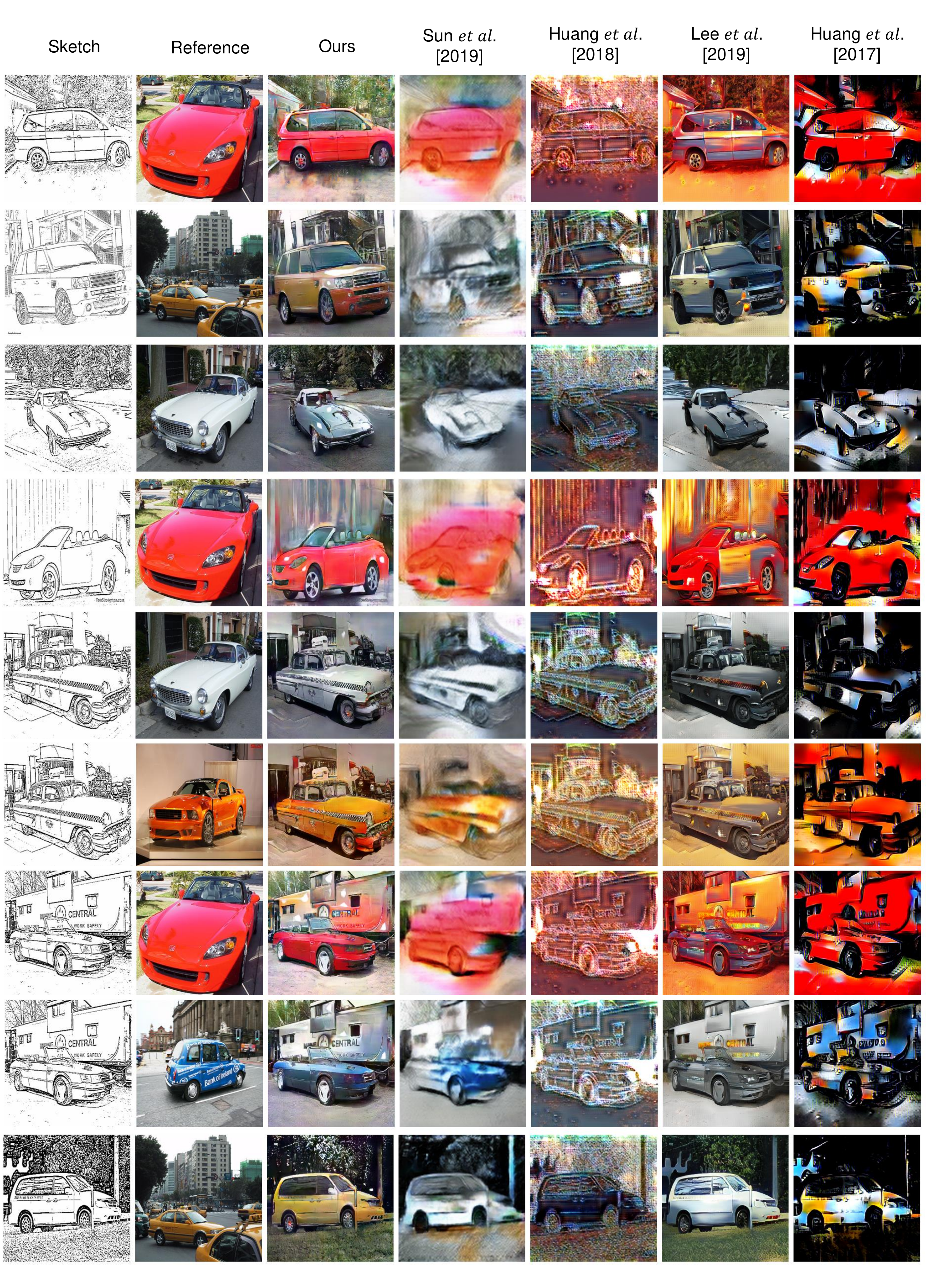}
\end{center}
   \caption{Qualitative comparisons with baselines on the ImageNet~\cite{russakovsky2015imagenet} dataset.}
\label{fig:supp_baselines_car}
\end{figure*}

\begin{figure*}
\begin{center}
\includegraphics[width=1\linewidth]{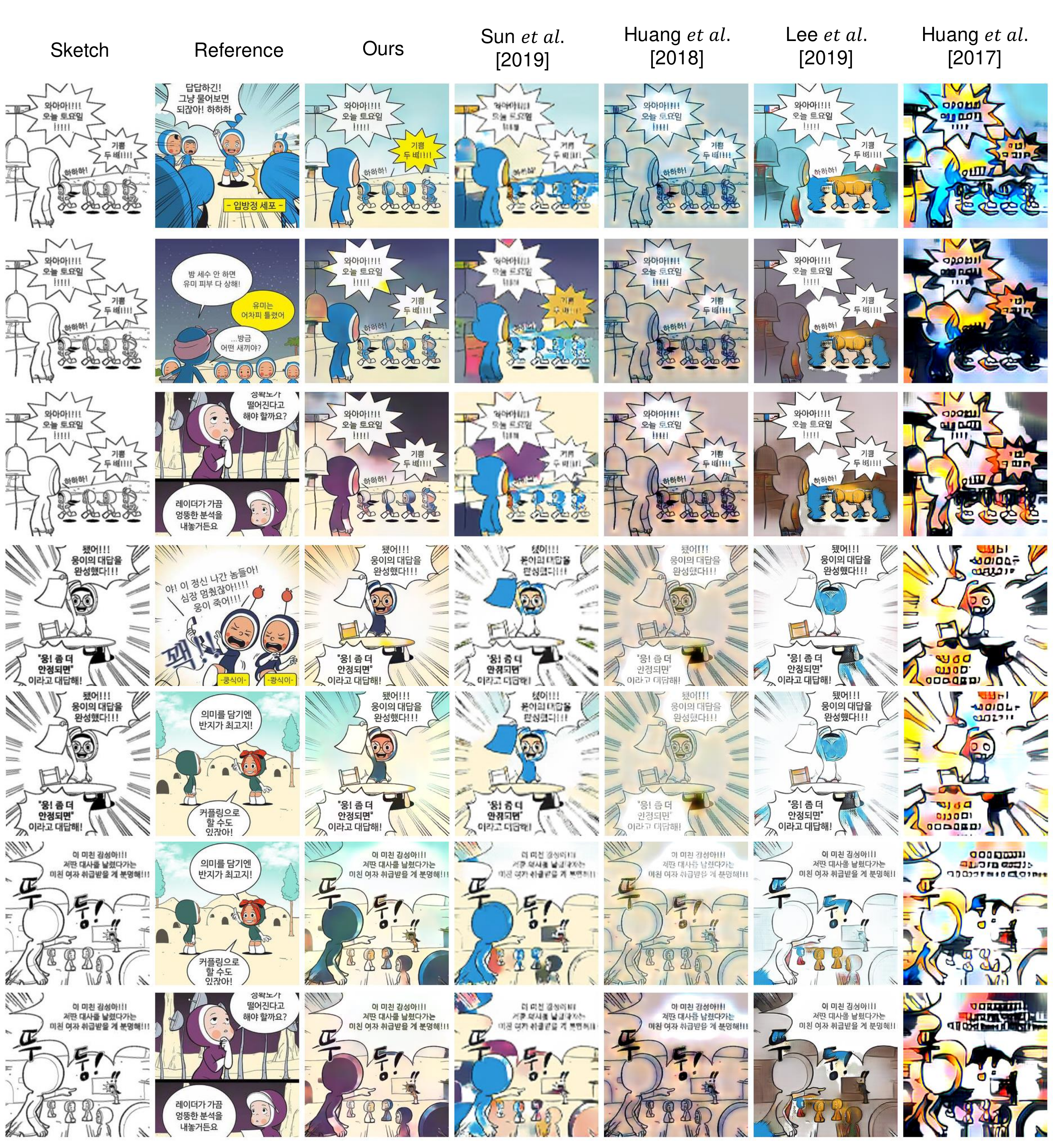}
\end{center}
   \caption{Qualitative comparisons with baselines on the Yumi's Cells~\cite{yumicells} dataset.}
\label{fig:supp_baselines_yumi}
\end{figure*}

\begin{figure*}
\begin{center}
\includegraphics[width=1\linewidth]{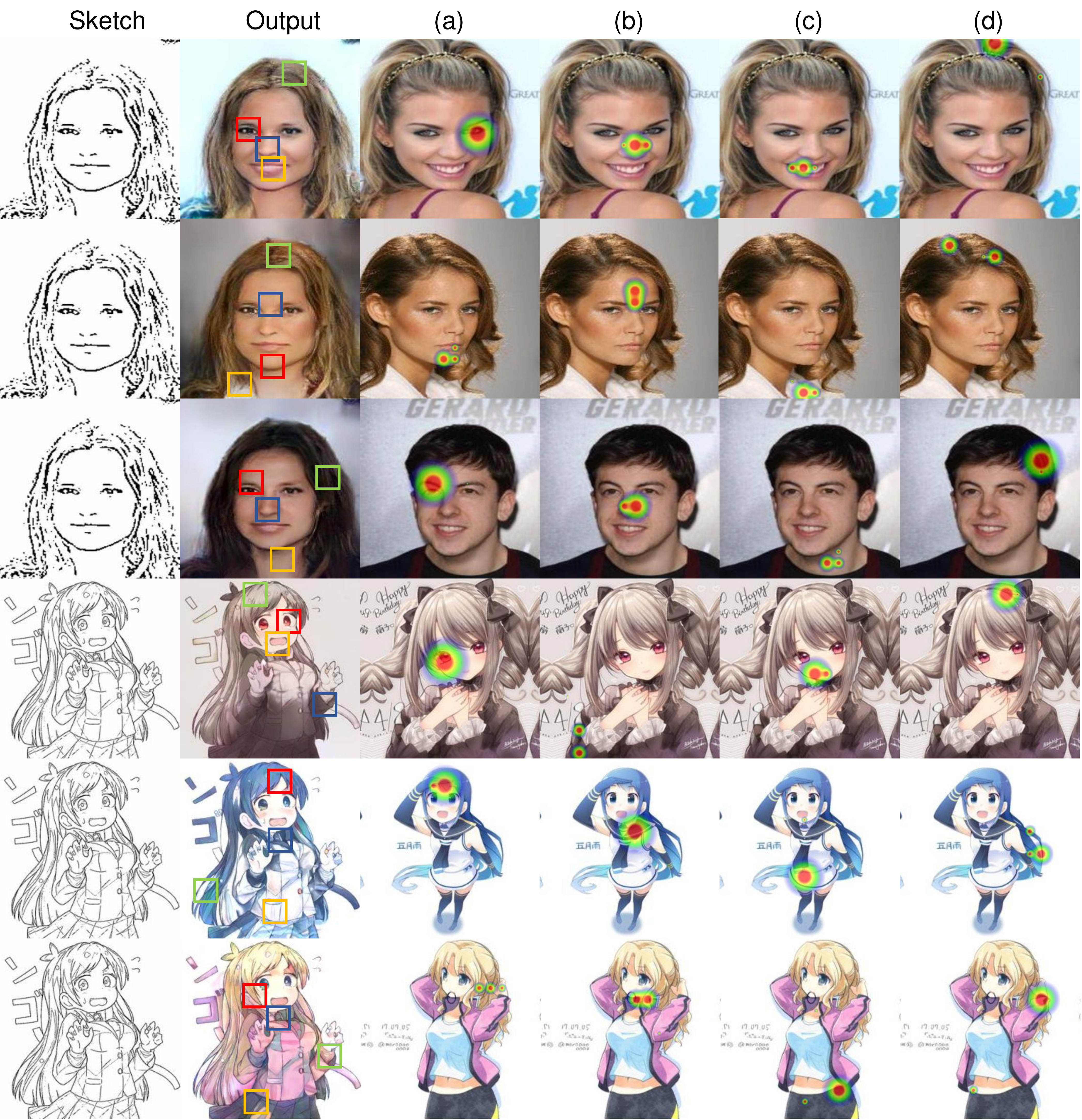}
\end{center}
  \caption{The visualization of attention maps on CelebA and Tag2pix dataset. The colored squares on the second column indicate the query region and corresponding key regions are highlighted in the next four columns. The different color of square means the different query region, and each red, blue, yellow, and green corresponds with the column (a), (b), (c), and (d), respectively.}
\label{fig:supp_attention_vis2}
\end{figure*}

\begin{figure*}
\begin{center}
\includegraphics[width=0.85\linewidth]{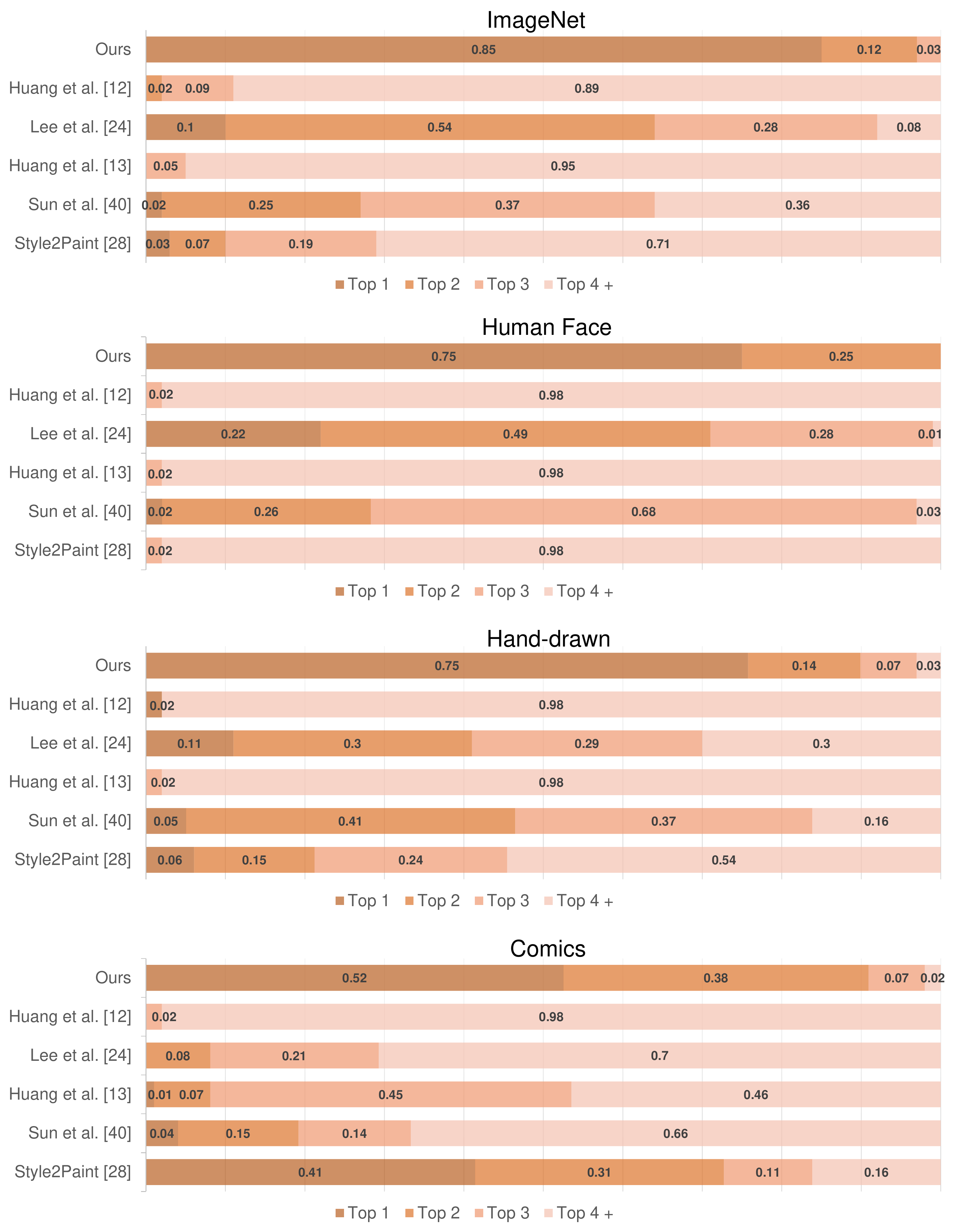}
\end{center}
  \caption{The results of the user study for comparison between our model and existing baselines. Question type 1: Overall Colorization Quality and Realism.}
\label{fig:subb_userstudy_1}
\end{figure*}

\begin{figure*}
\begin{center}
\includegraphics[width=0.85\linewidth]{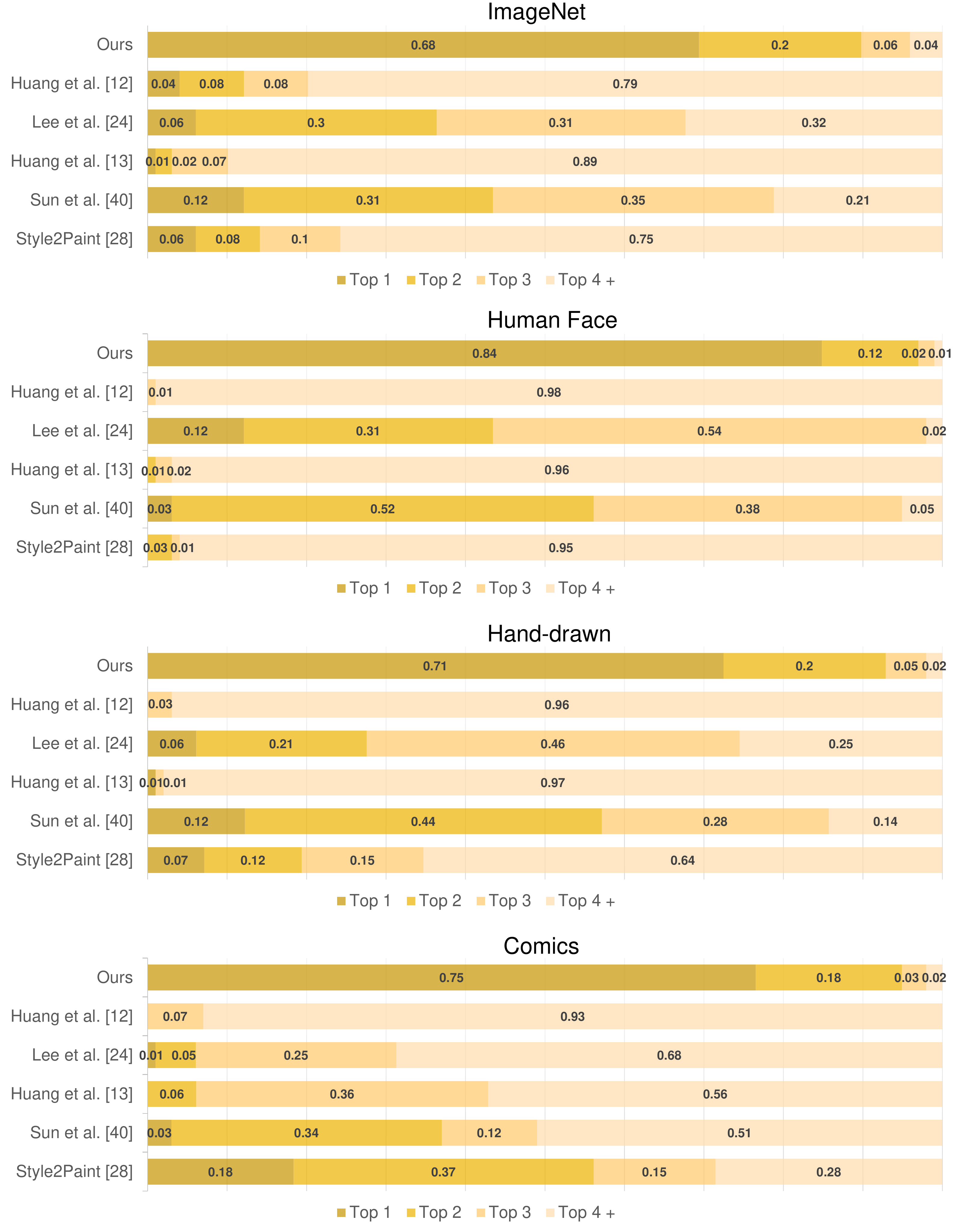}
\end{center}
  \caption{The results of the user study for comparison between our model and existing baselines. Question type 2: Detailed Reflection of Reference.}
\label{fig:subb_userstudy_2}
\end{figure*}

\begin{figure*}
\begin{center}
\includegraphics[width=0.85\linewidth]{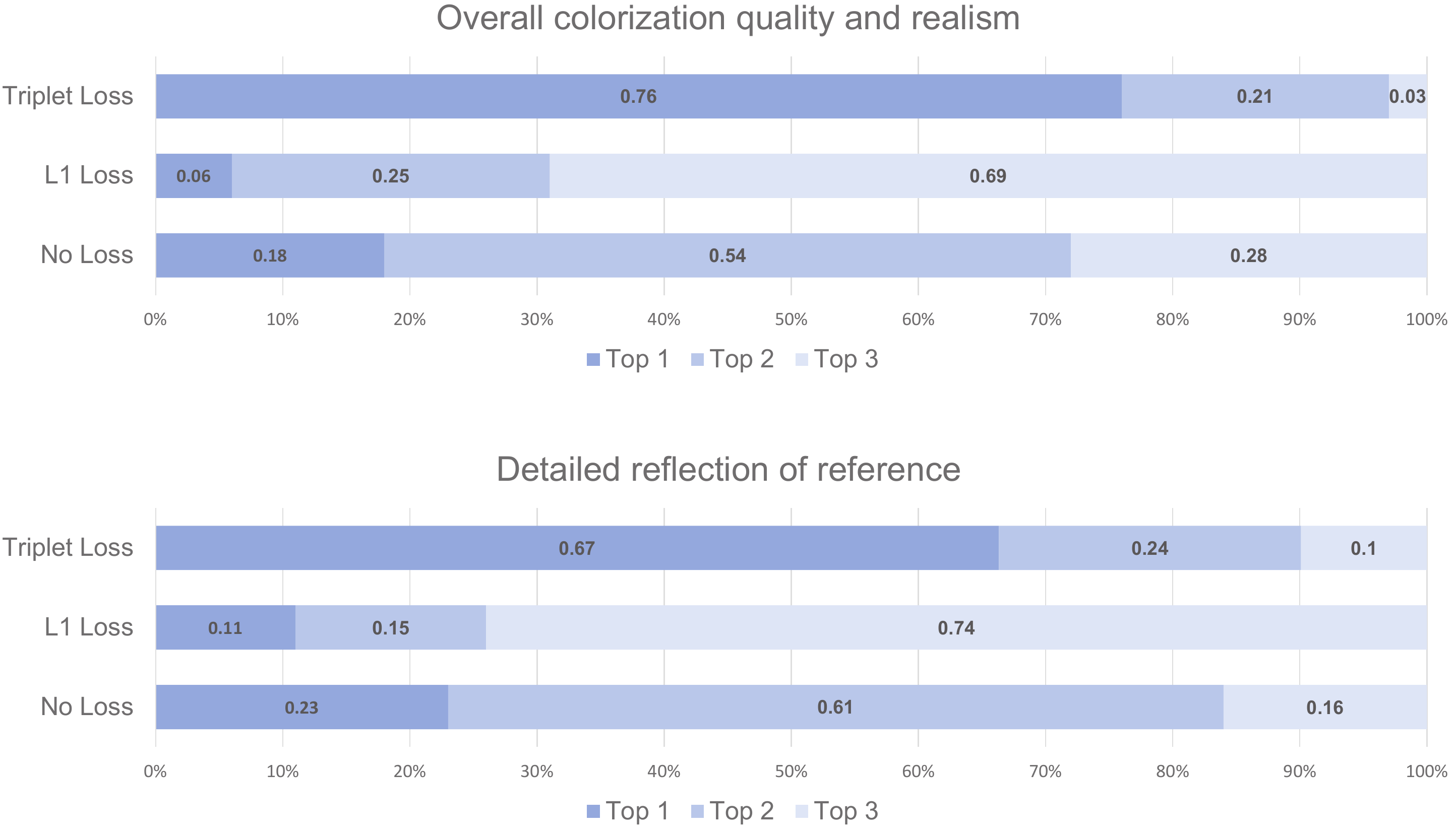}
\end{center}
  \caption{The results of the user study for comparison between model with triplet loss, $L_{1}$-loss and no loss. The percentages are averaged over all the datasets.}
\label{fig:subb_userstudy_ablation}
\end{figure*}

% {\small
% \bibliographystyle{ieee_fullname}
% \bibliography{main}
% }

%%supplementary

\end{document}